\documentclass{article} 

\usepackage{colm2024_conference}
\usepackage{titletoc}
\usepackage{hyperref}
\usepackage[page,header]{appendix}
\usepackage{lipsum}
\usepackage{microtype}
\usepackage{hyperref}
\usepackage{url}
\usepackage{booktabs}
\usepackage{graphicx}
\usepackage{multirow}
\usepackage{xspace}
\usepackage[breakable]{tcolorbox}
\usepackage{amsmath}
\usepackage{caption}
\usepackage{subcaption}
\usepackage{bbding}
\usepackage{multirow}

\usepackage{pgfplots}
\pgfplotsset{compat=1.17}

\newtcolorbox{myboxnote}[1][]{
  breakable,
  title=#1,
  colback=gray!4, 
  colframe=gray!90
}

\definecolor{openaigreen}{RGB}{85, 180, 129}
\newtcolorbox{commentprompt}{
  colback=openaigreen!15, 
  colframe=gray!75,
  boxrule=1pt,
  boxsep=0pt,
  left=6pt, 
  right=6pt,
  top=6pt, 
  bottom=6pt, 
  fontupper=\small
}

\newcommand{\BosphorusLogo}{\raisebox{-.4em}{\rlap{\raisebox{.8em}{\hspace{1.em}\scriptsize}}\includegraphics[height=1.5em]{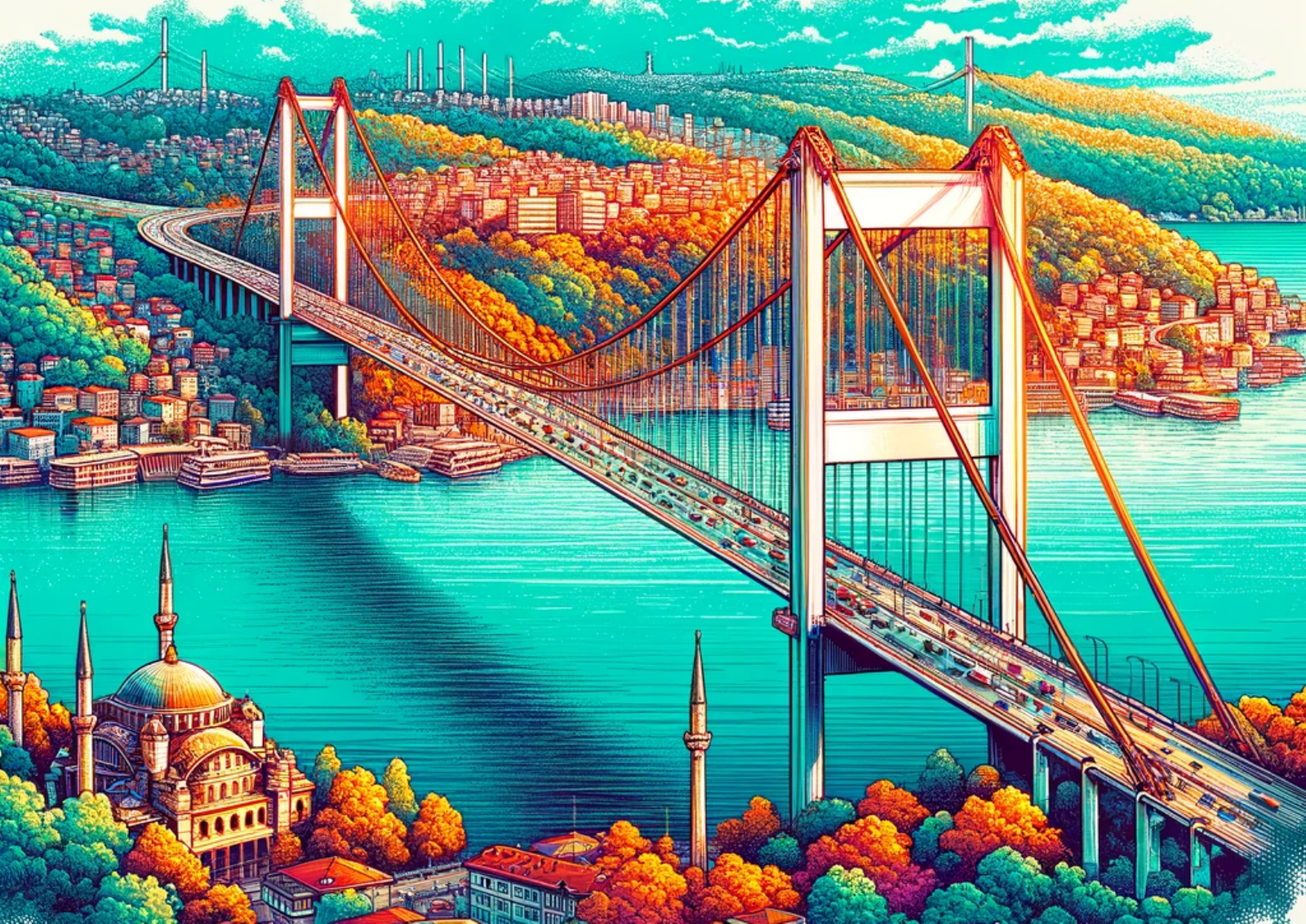}}\xspace}

\title{\BosphorusLogo~Bridging the Bosphorus: Advancing Turkish Large Language Models through Strategies for Low-Resource Language Adaptation and Benchmarking}

\author{
    Emre Can Acikgoz$^{1,2}$\thanks{Corresponding author, \href{mailto:eacikgoz17@ku.edu.tr}{eacikgoz17@ku.edu.tr}}\;,
    Mete Erdoğan$^{1,2}$\;,
    \textbf{Deniz Yuret}$^{1,2}$ \\[2mm]
    $^1$Koç University, KUIS AI Center, $^2$Koç University, Department of Computer Engineering\\[2mm]
}

\colmfinalcopy 
\begin{document}

\maketitle

\vspace{-6ex}
\begin{center}
     \url{https://emrecanacikgoz.github.io/Bridging-the-Bosphorus/}
\end{center}
\vspace{1ex}

\begin{abstract}
Large Language Models (LLMs) are becoming crucial across various fields, emphasizing the urgency for high-quality models in underrepresented languages. This study explores the unique challenges faced by low-resource languages, such as data scarcity, model selection, evaluation, and computational limitations, with a special focus on Turkish. We conduct an in-depth analysis to evaluate the impact of training strategies, model choices, and data availability on the performance of LLMs designed for underrepresented languages. Our approach includes two methodologies: (i) adapting existing LLMs originally pretrained in English to understand Turkish, and (ii) developing a model from the ground up using Turkish pretraining data, both supplemented with supervised fine-tuning on a novel Turkish instruction-tuning dataset aimed at enhancing reasoning capabilities. The relative performance of these methods is evaluated through the creation of a new leaderboard for Turkish LLMs, featuring benchmarks that assess different reasoning and knowledge skills. Furthermore, we conducted experiments on data and model scaling, both during pretraining and fine-tuning, simultaneously emphasizing the capacity for knowledge transfer across languages and addressing the challenges of catastrophic forgetting encountered during fine-tuning on a different language. Our goal is to offer a detailed guide for advancing the LLM framework in low-resource linguistic contexts, thereby making natural language processing (NLP) benefits more globally accessible.

\end{abstract}


\section{Introduction}
\label{sec: introduction}


The remarkable advancements in Large Language Models (LLMs) have revolutionized the field of natural language processing (NLP) \cite{gpt32020, gpt42023, palm2022, palm22023, llama22023}. However, addressing models that diverge from an English-centric framework poses considerable challenges, particularly in low-resource languages. While certain languages like Turkish aren't categorized as under-resourced, there's a limited number of research groups focusing on them \cite{mukayese}. Consequently, these languages lag in advancing cutting-edge systems because of the absence of solid and open-source base LLMs together with standardized benchmarks to evaluate their capabilities.


Recognizing this gap, our work is motivated by aiming to leverage Turkish LLMs. We meticulously demonstrate two distinct methodologies: we first tried to adapt two base LLMs, Mistral-7B \cite{mistral7b} and GPT2-xl \cite{gpt22019} to Turkish. Secondly, we trained a family of decoder models entirely from scratch in varying sizes. To adhere the Turkish LLMs to human instructions and extend their reasoning capabilities, we designed a novel Turkish instruction-tuning (IT) dataset, designed to enhance the reasoning abilities of Turkish LLMs by following the Self-Instruct framework \cite{selfinstruct}. 

One of the key challenges with Turkish LLMs is evaluating their accuracy on different tasks in a reproducible and fair manner while ensuring dataset quality. Many reasoning datasets have been directly machine-translated from English without any validation, leading to biased and inaccurate results. To address this, we introduce two Turkish datasets: TruthfulQA-TR, for assessing a model's tendency to reproduce common falsehoods, and ARC-TR, a set of grade-school science questions. We carefully translated by using state-of-the-art tools and validated all samples with multiple annotators, cleaning them as needed. We detailed the translation and annotation processes.

Our contributions are as follows:
\begin{itemize}
    \item We release the Hamza LLM series, encompassing models from 124M to 1.3B parameters. Notably, Hamza-xl with 1.3B parameters marks the premier and most expansive open-source, scientifically vetted Turkish LLM that is trained on 300B tokens.
    \item Our analysis explores two distinct methodologies for developing Turkish LLMs in resource and computational power-constrained environments: (i) extending pretrained models (Mistral-7b and GPT2-xl) with Turkish-only data (called as Hamza$_{\scriptsize Mistral}$ and Hamza$_{\scriptsize GPT2-xl}$), and (ii) constructing a model from scratch, similar to the GPT2 approach. This paper thoroughly discusses the merits and drawbacks of these strategies.
    \item We have curated new Turkish evaluation datasets TruthfulQA-TR and ARC-TR by carefully validating each with multiple annotators, offering meticulously cleaned datasets, and launching a leaderboard to catalyze ongoing advancements in Turkish LLMs.
    \item Committing to open science principles, we make all source codes, model checkpoints, and datasets open-source and publicly accessible.
\end{itemize}

By detailing the development of specialized datasets and methodologies, we offer a comprehensive guide for building LLMs for languages with limited resources. Additionally, our contributions substantially enrich the field by providing critical resources that will support future research in Turkish language processing and the broader area of Natural Language Processing (NLP) for under-resourced languages.

\section{Datasets}
\label{sec: datasets}

The initial step in building a base LLM involves pretraining it on a vast corpus of text with a next-token-prediction objective \cite{gpt32020}. This corpus comprises trillions of words gathered from the internet and is characterized by its large volume but often compromised in quality due to the noise in the raw internet data. Following the pretraining, the model undergoes fine-tuning with high-quality prompt-response pairs which focuses on improving the model's reasoning capabilities \cite{sftSurvey}. In the end, the goal is to achieve a Supervised-Finetuned (SFT) model that is aligned with the desired response behavior or domain expertise. This section describes the corpora utilized in the pretraining phase (Section \ref{sec: pretraining-dataset}) and the development process of the Turkish IT dataset (Section \ref{sec: it-dataset}).

\subsection{Pretraining Dataset}
\label{sec: pretraining-dataset}
For pretraining our models, we utilized CulturaX \cite{nguyen2023culturax}, a substantial multilingual dataset designed for LLM development. This dataset contains 6.3 trillion tokens in 167 languages and is a combination of two well-known multilingual datasets: mC4 \cite{mc4} and Oscar \cite{oscar1, oscar2, oscar3, oscar4, oscar5}. These datasets go through a detailed preprocessing that involves removing duplications, filtering out URLs, identifying languages, metric-based cleaning, and refining documents to enhance the data quality and consistency of each corpus. Since our focus is building a Turkish LLM, we only used the Turkish splits from CulturaX.

\paragraph{mC4.} mC4 \cite{mc4} is a large multilingual dataset initially created for training the mT5 \cite{mt5} which multilingual encoder-decoder model pretrained on 101 different languages. This dataset was generated by extracting content from 71 monthly snapshots of the internet via Common Crawl (CC). CulturaX contains version 3.1.0 of mC4\footnote{\url{https://huggingface.co/datasets/mc4}} which was provided by AllenAI. Its raw dataset contains 337GB of Turkish data.

\paragraph{OSCAR.} OSCAR (Open Super-large Crawled Aggregated coRpus) is a web-based multilingual dataset that is specialized in offering large volumes of unannotated raw data commonly used for training large deep learning models. It was developed by efficient data pipelines to organize and filter web data effectively. The final version of Oscar23\footnote{\url{https://huggingface.co/datasets/oscar-corpus/OSCAR-2301}} contains 73.7GB of Turkish data.

\paragraph{CulturaX Turkish.} We trained using the Turkish subset of CulturaX\footnote{\url{https://huggingface.co/datasets/uonlp/CulturaX}}, comprising 128 documents totaling 180 GB. No additional preprocessing was required since CulturaX had already undergone thorough detailed preprocessing during its creation. In the end, our Turkish dataset corpus comprises 130B unique tokens determined by using the GPT-2 tokenizer \cite{gpt22019}.

\begin{table*}[h]
\centering
\begin{tabular}{lrrr}
\toprule
\textbf{Corpus}     &\textbf{Documents}     &\textbf{Ratio}    &\textbf{\# of Tokens }     \\ \midrule
mC4                 & 75,859,899            & 80.52\%          & 104.3 B                          \\
OSCAR-2019          & 5,867,831             & 6.23\%           & 8.1 B                          \\
OSCAR-2109          & 6,614,512             & 7.02\%           & 9.1 B                          \\
OSCAR-2201          & 2,580,896             & 2.74\%           & 3.5 B                          \\
OSCAR-2301          & 3,284,322             & 3.49\%           & 4.5 B                          \\ \midrule
CulturaX (total)    & 94,207,460            & 100.0\%          & 129.5 B                          \\ \bottomrule
\end{tabular}

\caption{\textbf{Statistics of the pretraining dataset.} This table presents the statistics of our pretraining dataset used to train our Hamza series models that are presented in Table \ref{tab: archs}. }
\label{tab: datasets}
\end{table*}
\begin{table*}[h]
\centering
\begin{tabular}{lrrr}
\toprule
\textbf{Corpus Split}     &\textbf{Documents}     &\textbf{Portion}   &\textbf{\# of Tokens}     \\ \midrule
CulturaX 0.1GB            & 36,799                & 0.05\%          & 0.05 B                                \\
CulturaX 0.25GB           & 91,998                & 0.14\%          & 0.13 B                                \\
CulturaX 0.5GB            & 183,996               & 0.28\%          & 0.25 B                               \\
CulturaX 1.0GB            & 367,993               & 0.56\%          & 0.5 B                               \\ 
CulturaX 2.0GB            & 735,987               & 1.11\%          & 1.1 B                             \\ 
CulturaX 5.0GB            & 1,839,968             & 2.78\%          & 2.5 B                              \\ \bottomrule
\end{tabular}      
\caption{\textbf{Statistics of the continued pre-training dataset.} This table presents the statistics of our continued pretraining dataset that is used to train Hamza$_{\scriptsize Mistral}$ and Hamza$_{\scriptsize GPT2-xl}$.}
\label{tab: continued-pretraining-datasets}
\end{table*}


\subsection{Instruction-Tuning Dataset}
\label{sec: it-dataset}
Instruction fine-tuning is a crucial method used to improve LLMs in terms of their performance and ability to follow specific instructions \cite{it-survey}. 
This phase involves supervised training of LLMs using an instruction-tuning (IT) dataset composed of instruction and response pairs that link input instructions to their corresponding response outputs. 

\paragraph{Self-Instruct.} To create an automated, high-quality, and diverse IT dataset, we adapt the Self-Instruct procedure \cite{self-instruct, alpaca} for Turkish. 
We established 175 diverse instruction and response pairs as seed tasks which are translated manually from Alpaca repository\footnote{Alpaca Repository: \url{https://github.com/tatsu-lab/stanford_alpaca}} by human annotators. These annotators are experts in NLP and native speakers of both Turkish and English. 
For the given detailed prompt, we asked text-davinci3 \cite{gpt32020} to generate 20 complex and diverse instruction-response pairs, adhering strictly to the guidelines specified in the prompt. An example of a prompt is illustrated in Appendix \ref{sec: it-prompt}. Generated pairs are post-processed by removing any samples that contain visual context like images or photographs. This process resulted in the creation of 50,817 samples and cost only 8.12\$, which were then utilized for supervised fine-tuning (SFT).
 
\section{Methodology}
\label{sec: methodology}
Creating an LLM for under-resourced languages, like Turkish, often poses challenges primarily due to the scarcity of publicly available data especially if you have limited computational resources. Regarding these, we followed two different strategies to build a Turkish series of LLMs: (i) further training state-of-the-art base models on Turkish data, which was initially unfamiliar with Turkish (i.e., not trained on Turkish data), (ii) pretraining a model from scratch, following GPT2 scales on a vast amount of text data defined in Section \ref{sec: pretraining-dataset}.

\subsection{Method 1: Further Training a Base Model (Hamza$_{\scriptsize Mistral}$ and Hamza$_{\scriptsize GPT2-xl}$)}
\label{sec: method-1}
In this approach, we aim to enhance base LLMs with Turkish linguistic capabilities. After a detailed evaluation based on perplexity, we selected an LLM that did not specifically train on Turkish data during its initial pretraining phase. We subjected it to further training using Turkish-only data, accomplished through the next-token prediction objective implemented in an autoregressive manner. Essentially, this process can be regarded as a continuation of the pretraining phase of LLMs, but training on a specific portion of the Turkish dataset this time.

\paragraph{Selecting Base Model.} For the successful development of an advanced Turkish LLM with a 7 billion parameter scale, choosing the most suitable base model is essential. To this end, we have selected Mistral 7B \cite{mistral7b} as one of our base models, owing to its recent success across various tasks. Additionally, we opted for GPT2-xlarge, since our Hamza model is trained from scratch on the GPT2 architecture. This selection allows for a meaningful comparison between models trained from scratch and those initially trained in English and subsequently continued with pre-training in the same architectural setup.

\paragraph{Dataset.} In order to inject Turkish into Mistral and GPT-2 base LLMs, we followed a strategy of incremental continued pretraining on Turkish-specific segments of our dataset. Beginning with an initial 100MB of pure Turkish data, we progressively expanded the training corpus, culminating in the model being trained on 5GB of data. This volume aligns closely with the dataset size used for GPT \cite{gpt2018}, ensuring a comprehensive and effective adaptation of the model to handle Turkish linguistic nuances. Please refer to Table \ref{tab: continued-pretraining-datasets} for the details of these splits.

\paragraph{Training.} As a continual learning approach, we conducted a series of experiments by progressively enlarging the pretraining corpus size and halting upon observing convergence. The models are initialized with the pretraining weights of the Mistral-7B and GPT2-xl and then further trained on segments of our text corpus with a casual language modeling objective. Throughout our continued pretraining experiments, we employed LoRA \cite{lora2021} and updated only the additional bottleneck adapter weights while freezing the original model weights to make the training cost-efficient and avoid any catastrophic forgetting from the models' previous capabilities. During our LoRA trainings, we used $r=32$ and $\alpha=32$, along with a dropout rate of 0.05, applying LoRA exclusively to the projection layers. We used AdamW optimizer and cosine scheduler with a learning rate of $0.0001$. Based on our experiments, we opted for a batch size of 1 and avoided gradient accumulation due to its significant impact on convergence. To simplify the execution of our experiments and ensure the reproducibility of our results, we used the LLaMA-Factory\footnote{\url{https://github.com/hiyouga/LLaMA-Factory}} repository, only in our LoRA-based continued pretraining experiments.

\begin{table*}[h]
\resizebox{\linewidth}{!}{%
\begin{tabular}{lccccccc}
\toprule
\textbf{Model} & \textbf{Parameters} & \textbf{Layers} &  \textbf{Heads} & \textbf{d$_{\textbf{\scriptsize model}}$} & \textbf{Learning Rate} & \textbf{Batch Size}  & \textbf{Tokens} \\ \midrule
hamza-small    & 124M                & 12              & 12              & 768                                       & 6.0$e^{-4}$            & 0.5M                 & 300B            \\
hamza-medium   & 354M                & 24              & 16              & 1024                                      & 3.0$e^{-4}$            & 0.5M                 & 300B            \\
hamza-large    & 772M                & 36              & 20              & 1280                                      & 3.0$e^{-4}$            & 0.5M                 & 300B            \\
hamza-xlarge   & 1.3B                & 24              & 16              & 2048                                      & 2.0$e^{-4}$            & 0.5M                 & 300B            \\ \bottomrule
\end{tabular}
}
\caption{\textbf{Architecture and optimization hyperparameters for the 4 Hamza model sizes that trained from scratch.}}
\label{tab: archs}
\end{table*}

\subsection{Method 2: Pretraining from Scratch (Hamza Series Models)}
\label{sec: method-3}

In our final approach for developing a Turkish base-LLM, we adopted the most straightforward method: training from scratch using Turkish-only datasets. We follow a similar framework as in GPT2 \cite{gpt22019}, with similarities in training procedures and architectural settings. However, we differed in our approach by utilizing a pretraining corpus nearly double the size of GPT2. All our pretraining code is openly accessible\footnote{\href{https://github.com/emrecanacikgoz/turkish-llm}{https://github.com/emrecanacikgoz/turkish-llm}}.

\paragraph{Pretraining Data.} The construction of a robust LLM hinges on the aggregation and processing of high-quality text data. To develop Hamza, we used the Turkish split of CulturaX \cite{nguyen2023culturax} includes a meticulous process of data curation. It gathers a comprehensive dataset from open-sources mC4 \cite{mc4} and OSCAR \cite{oscar1, oscar2, oscar3, oscar4, oscar5}. Our pretraining data contains 128 parquet files each 1.4GB, totaling almost 179.2GB. The compiled training dataset contains 129,486,207,634 (130B) training tokens. Further details of the data gathering, structure, and preparation can be found in CulturaX work \citet{nguyen2023culturax}.

\paragraph{Architecture.} To develop an inaugural Turkish base model, we followed prior works, establishing a solid model for Turkish language modeling akin to earlier studies on other languages. Our approach led to the creation of four variants of Hamza, following GPT-2 \cite{gpt22019}: Hamza-small (124M parameters), Hamza-medium (354M parameters), Hamza-large (772M parameters), and our largest model, Hamza-xlarge (1.3B parameters). The architectural specifications of these models are given in Table \ref{tab: archs}.

\paragraph{Optimizer.} During our training, AdamW \cite{adamw} optimizer is used with hyper-parameters $\beta_1=0.9$ and $\beta_2=0.95$. A cosine learning rate schedule is implemented, designed to reduce the learning rate to 10\% of its maximum value. Additionally, we applied a weight decay rate of $0.1$ and limited the gradient norm to $1.0$ to prevent overfitting. The training process includes $2,000$ warm-up steps. We used a learning rate 0f $0.0006$ and batch size $491,520$ in our smallest model Hamza-small. We varied the learning rate and batch size according to the model size, for details see Table \ref{tab: archs}.

\paragraph{Training.} Our from-scratch Hamza models are built on the GPT2 architecture \cite{gpt22019} and incorporate the flash-attention mechanism for efficient training \cite{flashattention}. As outlined in Table \ref{table: hamza-configurations}, the hyperparameters of the model follow the scaling principles set by GPT2, except for the largest variant, Hamza-xlarge, which is inspired by a recent French-based LLM \cite{croissantllm}. All model versions were trained for 300 billion tokens, with a uniform batch size of 500,000 tokens. The learning rate was fine-tuned for each model variant. We standardized the context window across all models at 1024 tokens and did not employ any dropout techniques during their training process. All training sessions were conducted in half-precision (fp16) settings by utilizing both tensor and data parallelism across eight A100 GPUs each with 80GB of memory.

\begin{table*}[h]
\centering
\begin{tabular}{clccc}
\toprule
\textbf{Type} & \textbf{Models} & \multicolumn{2}{c}{\textbf{Accuracy ($\uparrow$)}} &\textbf{BPC ($\downarrow$)} \\ \cmidrule(lr){1-1}  \cmidrule(lr){2-2} \cmidrule(lr){3-4} \cmidrule(lr){5-5}
\textbf{}    &\textbf{}         &\textbf{ARC-TR}  & \textbf{TruthfulQA-TR}    & \textbf{\textbf{trnews-64}}   \\

\multirow{7}{*}{\shortstack[c]{\textbf{Base \& SFT} \\ \textbf{Models}}} 
& LLaMA2 7b               & 25.94          & 41.18                    & 1.374                 \\
& LLaMA3 8b               & 43.09          & 44.77                    & 0.929                 \\
& Mistral 7b              & 32.68          & 41.16                    & 1.260                 \\
& Gemma 2B                & 31.31          & 43.57                    & 1.208                 \\
& Gemma 7B                & \textbf{46.16} & 42.35                    & 0.989                 \\
& GPT2-xl                 & 24.91          & 40.97                    & 2.533                 \\ 
& LLaMA2 7b-chat          & 25.00          & 40.07                    & 1.374                 \\
& Mistral 7b-chat-v2      & 35.24          & \underline{48.34}        & 1.428                 \\ \midrule

\multirow{6}{*}{\shortstack[c]{\textbf{Multi-} \\ \textbf{lingual} \\ \textbf{Models}}} 
& XGLM-7.5B               & 29.01           & 39.09                    & 0.880                 \\
& XGLM-4.5B               & 25.94           & 40.18                    & 0.949                 \\
& XGLM-2.9B               & 27.05           & 39.35                    & 0.946                 \\ 
& XGLM-1.7B               & 26.37           & 41.75                    & 1.044                 \\ 
& XGLM-564M               & 23.55           & 42.59                    & 1.125                 \\ 
& mGPT                    & 26.54           & 42.37                    & 1.306                 \\ \midrule

\multirow{11}*{\shortstack[c]{\textbf{Huggingface} \\ \textbf{Turkish} \\ \textbf{Models}}} 
& Kanarya-2b              & 29.78                     & 41.43                    & \textbf{0.724}        \\
& Kanarya-750m            & 28.16                     & 41.50                    & 0.767                 \\
& Turkcell-LLM-7b-v1      & 43.09                     & 44.91                    & 1.208                \\
& ytu-gpt2-large          & 27.13                     & 43.09                    & 0.805                 \\
& Trendyol-7b-base        & 35.24                     & 41.50                    & 0.829                 \\ 
& Trendyol-7b-chat        & 35.58                     & 44.35                    & 0.820                 \\ 
& Trendyol-7b-dpo         & 39.93                     & \textbf{50.11}           & 0.859                 \\
& Commencis-LLM           & 33.28                     & 44.50                    & 1.306                 \\
& Sambalingo-tr           & \underline{44.37}         & 46.61                    & 0.894                 \\
& Thestral-tr-chat        & 34.00                     & 41.90                    & 1.314                 \\ 
& Mistral-7b-chat-v2-tr   & 33.96                     & 45.71                    & 1.411                 \\ 
& Gemma-2B-tr             & 31.31                     & 44.46                    & 1.089                 \\ \midrule

\multirow{6}{*}{\textbf{\shortstack[c]{\textbf{Our} \\ \textbf{Models}}}} 
& Hamza-small                    & 25.26                       & 43.65                    & 0.897                 \\
& Hamza-medium                   & 26.45                       & 43.55                    & 0.814                 \\
& Hamza-large                    & 29.10                       & 40.93                    & 0.760                 \\
& Hamza-xl                       & 28.24                       & 42.33                    & \underline{0.754}     \\
& Hamza$_{\scriptsize GPT2-xl}$  & 24.74                       & 44.95                    & 1.152                 \\  
& Hamza$_{\scriptsize Mistral}$  & 39.85                       & 46.40                    & 0.816                 \\ \bottomrule
\end{tabular}

\caption{\textbf{Performance comparison on various Turkish tasks.} We compare the performance of various types of models: (i) Base and SFT Models, (ii) Multilingual Models (iii) Open-Source HF (Huggingface) Turkish Models, (iv) Our pretrained and adapted Hamza Models. The first two columns show accuracies evaluated on the ARC-TR and TruthfulQA-TR datasets. The last column includes the Bits-Per-Character (BPC) metric evaluated on \textsc{trnews-64} corpus. Note that Accuracy is the highest and BPC is the lowest for the best models. The top-performing model for each metric is highlighted in bold, while the second-best model is underlined for easy identification. See Appendix \ref{sec: models} for model details.}
\label{tab: acc-results}
\end{table*}



\section{Evaluations}
\label{sec: evaluation}


\subsection{Bits-Per-Character (BPC) Evaluations}
\label{sect: bpceval}
Auto-regressive language modeling is trained on optimizing the Negative Log-Likelihood (NLL) of the data in the training set and the effectiveness of the model is then calculated on the unseen test data. Furthermore, the most common metric to evaluate these models is perplexity, which measures the uncertainty of an LLM in predicting the next token in a sequence and is derived by taking the exponential average of the NLL. However, as various tokenizers can divide each sentence into differing numbers of tokens, NLL and PPL may produce incomparable results for models utilizing different tokenizers. To tackle this, we use Bits-Per-Character (BPC), which is another critical metric derived from NLL, used for evaluating the performance of LLMs at character-level. Further details on the calculation of these metrics are given in the Appendix in Section \ref{sect: bpceval}. Consequently, our comparisons mainly relied on BPC, which normalizes the impact of tokenization differences. For the BPC evaluation, we utilized the test set of the trnews-64 corpus \cite{mukayese}, comprising 5,000 samples.

\paragraph{Results.} We present the BPC results of different models evaluated on trnews-64 in the last column of Table \ref{tab: acc-results}; including our models together with various open-source multi-lingual and Turkish LLMs. Looking at the BPC results, we observe a wide range of values across the models. Lower BPC values indicate better performance in terms of compression, suggesting that the model is more efficient in representing the text. The most favorable outcomes are attained with the pretrained Kanarya-2b and Hamza-xlarge models. The adapted models which are originally pretrained on English but extended to Turkish, yielded promising results as well, lower than 1 BPC, whereas the multilingual models had a relatively lower performance.

\subsection{Prompting \& Few-Shot}
Evaluating the reasoning capabilities of LLMs in downstream Question Answering (QA) tasks is crucial to assess their abilities. However, finding such datasets in languages other than English is challenging due to the limited availability of benchmarks in these languages. To bridge this gap, we developed TruthfulQA-TR and ARC-TR Turkish question-answering datasets, that are designed to evaluate the ability of LLMs to generate truthful and accurate responses to questions. To develop the Turkish version of the main TruthfulQA Multiple Choice (MC) \cite{truthfulqa} Dataset and ARC (AI2 Reasoning Challenge) \cite{allenai:arc} dataset, we translated each example of these datasets using the advanced DeepL Machine Translation (MT) framework by its Python-supported API\footnote{\url{https://github.com/DeepLcom/deepl-python}}. After translating to Turkish, each sample was reviewed for errors or superficial translations. We used the test sets from TruthfulQA-MC2 and ARC-Challenge for evaluations. For more details on datasets and annotation validation, see the Appendix in Section \ref{sec: validation}. Our experiments followed the same prompting settings with LLM-Leaderboard\footnote{\url{https://huggingface.co/spaces/HuggingFaceH4/open_llm_leaderboard}}. We include performances of all of our models together with all the open-source Turkish LLMs that are available on Huggingface, together with other monolingual and multilingual models.

\paragraph{Results.} We evaluate various language models in depth, including base LLMs \cite{llama22023, mistral7b, gpt22019}, multilingual LLMs \cite{mgpt, xglm}, all available Turkish fine-tuned LLMs on Huggingface, and the models we propose in this paper. Our evaluation was conducted on the newly established Turkish Benchmarks, ARC-TR, in 25-shot settings, as well as on TruthfulQA-TR, adhering to the same settings as outlined by the LLM-Leaderboard. In ARC-TR, Google's Gemma 7B model leads with an accuracy of 46.16 even though it is not specifically tuned for Turkish, closely followed by Sambalingo-tr with 44.37 accuracy. Moreover, in the TruthfulQA-TR evaluation, Trendyol's DPO model emerges as the top performer with an accuracy of 50.11, while Mistral-7b-chat-v2 secures the second position with 48.34 accuracy. The accuracy scores for ARC-TR range from 24 to 47, and for TruthfulQA-TR, from 33 to 50. These results underscore the necessity for substantial improvements in these models to reach the proficiency levels observed in English benchmarks.

\paragraph{Qualitative Analysis.} We performed qualitative analysis on our models by testing them with various prompts as demonstrated in Appendix in Section \ref{sec: sample-model-outputs}. Both the pretrained and Hamza models perform well on sentence completion. In particular, compared to other open-source Turkish models, we observed a reduced tendency to generate text that resembles web-based content, where most of the Turkish corpora is retrieved from websites. Furthermore, we tested our models on English prompts to assess their ability to handle multilingual tasks. Although results indicate that our models can generate coherent responses, there is a high tendency for the models to continue English sentences in Turkish. Overall, our qualitative analysis highlights the robust performance of our models and the potential for diverse tasks.
\section{Case Studies}
\label{sec: case-studies}

\subsection{Enhancing Non-English Models: Fine-Tuning vs. From-Scratch Training}
The analysis of Turkish language models, specifically comparing models trained from scratch, continued pretraining from GPT2-xl \cite{gpt22019}, and those adapted using Mistral 7B \cite{mistral7b}, shows insightful trends. According to Table \ref{tab: case-studies-4-1}, the Mistral 7B adapted model exhibits superior performance on Turkish question-answering tasks, compared to other methods. Moreover, starting from scratch surpasses the continued pretraining approach within the same model architecture, underscoring the significance of the base language model when undertaking continued pretraining. This is evidenced by the discrepancy in accuracy between models fine-tuned from Mistral 7B versus those from GPT2. Therefore, applying continued pretraining to a robust base language model emerges as the most effective strategy for low-resource languages, considering both data scarcity and hardware constraints.
\begin{table*}[h!]
\centering
\begin{tabular}{lccc}
\toprule
\textbf{Models}         &\textbf{ARC-TR}     &\textbf{TruthfulQA-TR}    &\textbf{Avg.}  \\ \midrule
Hamza-xl                       & 28.24                       & 42.33                    & 35.28     \\
Hamza$_{\scriptsize GPT2-xl}$  & 24.74                       & 44.95                    & 34.84                 \\  
Hamza$_{\scriptsize Mistral}$  & \textbf{39.85}              & \textbf{46.40}                    & \textbf{43.12}                 \\ \bottomrule
\end{tabular}

\caption{\textbf{Accuracy comparison of our best models on Turkish question answering tasks.} This table shows the performance of our models, pretrained Hamza models with different sizes, and the Hamza$_{\scriptsize Mistral}$ and Hamza$_{\scriptsize GPT2-xl}$ models that are adapted on Turkish. We present the results evaluated on the ARC-TR (25 shot) and TruthfulQA-TR (6 shot) datasets.}
\label{tab: case-studies-4-1}
\end{table*}

\subsection{Effect of Supervised Fine-Tuning: Assessing Model Performance with the Proposed IT Dataset.}
Supervised Fine-Tuning (SFT) plays a crucial role in enhancing the reasoning capabilities of LLMs, as highlighted in existing research \cite{sftSurvey}. In this context, we introduced a novel Turkish IT Dataset, meticulously crafted from the ground up, inspired by the Alpaca \cite{alpaca, self-instruct}. By fine-tuning our largest model Hamza-xlarge with this bespoke Turkish IT Dataset, we observed an improvement in model performance across downstream benchmarks (see Table \ref{tab: case-studies-4-1}). This improvement underscores the effectiveness of SFT when applied to our tailored IT dataset, bolstering our model's reasoning proficiency slightly.
\begin{table*}[h!]
\centering
\begin{tabular}{lccc}
\toprule
\textbf{Models}       &\textbf{ARC-TR}                &\textbf{TruthfulQA-TR}    &\textbf{Avg.}   \\ \midrule 
Hamza-xl              & 28.24                         & 42.33                    & 35.28     \\
Hamza-xl + SFT        & 29.61                         & 44.67                    & 37.14           \\\bottomrule
\end{tabular}

\caption{\textbf{Accuracy results of our models fine-tuned on our Self-Instruct IT dataset on Turkish question answering tasks.} This table compares the performance increase after instruction tuning with dataset IT dataset described in Section \ref{sec: it-dataset}. We present the results evaluated on the ARC-TR (25 shot) and TruthfulQA-TR (6 shot) datasets.}
\label{tab: case-studies-4-2}
\end{table*}

\subsection{Retention after Fine-Tuning: Will Models Forget English-Learned Skills When Fine-Tuning on Another Language?}

\begin{figure}[htbp]
    \begin{minipage}[h]{0.56\textwidth}
    \scalebox{0.85}{
        \begin{tabular}{lcccc}
        \toprule
        \textbf{Models}                            &\textbf{ARC}                        &\textbf{TruthfulQA}       &\textbf{Avg.}   \\ \midrule
        GPT2-xl                                    & 30.29                              & 38.53                    & 34.41           \\ 
        Hamza$_{\scriptsize GPT2-xl}$ (0.1GB)      & 28.84                              & 38.15                    & 32.98           \\
        Hamza$_{\scriptsize GPT2-xl}$ (0.25GB)     & 26.37                              & 38.10                     & 32.88           \\
        Hamza$_{\scriptsize GPT2-xl}$ (0.5GB)      & 27.13                              & 38.88                     & 33.35           \\
        Hamza$_{\scriptsize GPT2-xl}$ (1GB)        & 26.54                              & 38.95                     & 33.09           \\
        Hamza$_{\scriptsize GPT2-xl}$ (2GB)        & 24.74                              & 40.34                     & 33.01           \\
        Hamza$_{\scriptsize GPT2-xl}$ (5GB)        & 22.61                              & 41.36                     & 32.49           \\ \midrule
        Mistral-7b                                 & 61.52                              & 42.57                   & 51.49           \\ 
        Hamza$_{\scriptsize Mistral}$ (0.1GB)      & 56.14                              & 40.31                   & 48.22           \\
        Hamza$_{\scriptsize Mistral}$ (0.25GB)     & 52.90                              & 39.15                   & 45.77           \\
        Hamza$_{\scriptsize Mistral}$ (0.5GB)      & 52.39                              & 38.70                   & 45.51           \\
        Hamza$_{\scriptsize Mistral}$ (1GB)        & 51.71                              & 41.46                    & 46.60           \\
        Hamza$_{\scriptsize Mistral}$ (2GB)        & 49.32                              & 38.44                   & 43.91           \\
        Hamza$_{\scriptsize Mistral}$ (5GB)        & 45.90                              & 40.90                   & 43.82           \\ \bottomrule
        \end{tabular}
        }
        \label{tab: case-studies-4-4x}
    \end{minipage}
    \hfill
    \begin{minipage}[c]{0.78\textwidth}
        \scalebox{0.75}{
        \begin{tikzpicture}
        \begin{axis}[
            title={\textbf{Comparison of ARC Scores of Mistral Model}},
            xlabel={\textbf{\# of Turkish Tokens}},
            ylabel={\textbf{\textcolor{red}{English} \&  \textcolor{blue}{Turkish} ARC Acc.}},
            xtick={0,1,2,3,4,5,6},
            xticklabels={
                0,
                50M,
                130M,
                250M,
                0.5B,
                1.1B,
                2.5B,
            },
            xlabel style={at={(0.5,-0.10)}},
            legend pos=outer north east,
            ymajorgrids=true,
            grid style=dashed,
            legend style={cells={align=left}},
            width=0.75\textwidth,
            xtick distance=5,
            legend style={at={(0.97,0.88)},anchor=east},
        ]
        
        \addplot[
            color=blue,
            mark=o,
            thick,
            nodes near coords,
            point meta=explicit symbolic,
        ] coordinates {
            (0,32.68)
            (1,33.70)
            (2,36.86)
            (3,36.01)
            (4,38.57)
            (5,38.91)
            (6,39.85)
        };
        
        \addplot[
            color=red,
            mark=square,
            thick,
            nodes near coords,
            point meta=explicit symbolic,
        ] coordinates {
            (0,61.52)
            (1,56.14)
            (2,52.90)
            (3,52.39)
            (4,51.71)
            (5,49.32)
            (6,45.90)
        };
        
        \legend{ARC-tr Acc., ARC-eng Acc.}
        \end{axis}
        \end{tikzpicture}
        }
    \end{minipage}
\caption{\textbf{Accuracy comparison of Continued Pretrained models on English (Left, Right) and
Turkish (Right) question answering tasks and demonstrating the original language catastrophic forgetting while learning the new language.} In the table on the left, the performance of our Hamza$_{\scriptsize Mistral}$ and Hamza$_{\scriptsize GPT2-xl}$ models that are adapted on Turkish together with the original Mistral 7B and GPT2-xl. We present the result of our ablation study, where the performance of the adapted models is given by progressively enlarging the pretraining corpus size from 0.1 GB to 5 GB. Here, the zero and few-show accuracies were evaluated on the original ARC and TruthfulQA. The figure on the right illustrates the Mistral model's results on both Turkish and English versions of the ARC dataset, highlighting its improved performance in Turkish and decreasing performance in English with continued pretraining. }
\label{tab: case-3-new}
\end{figure}
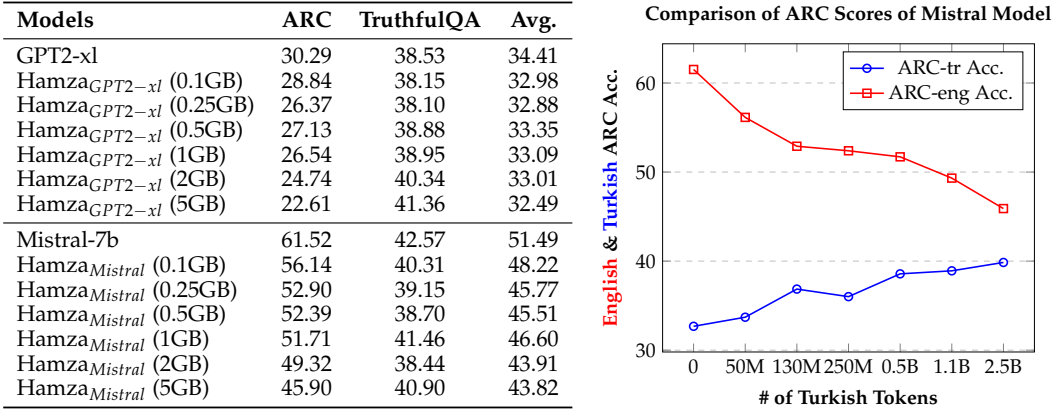

According to Table \ref{tab: case-3-new}, further pretraining of base English language models such as GPT2 and Mistral results in a decrease in accuracy proportional to the number of samples used during continued pretraining on the English downstream tasks TruthfulQA and ARC, compared to their original base scores before fine-tuning on Turkish. This indicates \textit{catastrophic forgetting}, where the models lose their prior knowledge upon being fine-tuned on a smaller language dataset, as evidenced by a decline in baseline accuracy compared to the versions not previously trained, even after applying techniques like LoRA training. One further work for this could be including some English data along with Turkish in each batch during continued pretraining.

\section{Conclusion}
\label{sec: conclusion}

Our work advances the development of Turkish LLMs, presenting a new series of models both trained from scratch (Hamza) and also adapted from other base LLMs (Hamza$_{\scriptsize Mistral}$ and Hamza$_{\scriptsize GPT2-xl}$), together with new Instruction Tuning dataset and a meticulously crafted Turkish LLM Leaderboard. In our analysis, we noted that the base LLMs exhibited catastrophic forgetting of their primary language knowledge during continued pretraining. Additionally, through the creation of a novel Turkish LLM evaluation benchmark, we have identified a significant performance gap between current Turkish LLMs and their English counterparts, underscoring the need for further improvements in Turkish language modeling. For more detailed discussions on limitations and future work, please refer to Appendix \ref{sec: limitations-and-future-work}. Our fully open-source work and detailed observations play a pivotal role in the field of Turkish language modeling, providing insights on construction methodologies and offering a comparative framework for evaluating performance, thereby paving the way for future advancements.

\newpage
\section*{Acknowledgments}
We extend our heartfelt gratitude to Kerem Zaman and Abdullatif Köksal for their exceptional support when needed. We also thank Ekin Akyürek for his insightful advice and in-depth discussions. Also, thanks to Ali Safaya for his assistance in organizing the benchmark annotation process. Additionally, we are grateful to all the dedicated annotators from Koç University students. On the other hand, this work is supported in part provided by the KUIS AI Center. Finally, special thanks to HyperBee AI for their help during model evaluations. The numerical calculations reported in this paper were fully/partially performed at TUBITAK ULAKBIM, High Performance and Grid Computing Center (TRUBA resources). Last but not least, we also acknowledge VSB – Technical University of Ostrava, IT4Innovations National Supercomputing Center, Czech Republic, for awarding this project access to the LUMI supercomputer, owned by the EuroHPC Joint Undertaking, hosted by CSC (Finland) and the LUMI consortium through the Ministry of Education, Youth and Sports of the Czech Republic through the e-INFRA CZ.

\bibliography{colm2024_conference}
\bibliographystyle{colm2024_conference}

\newpage
\section*{Appendix}
\appendix
\startcontents[section]
\printcontents[section]{l}{0}{\setcounter{tocdepth}{1}}
\newpage

\section{Related Work}
\label{sec: related-work}
\vspace{-0.25em}

LLMs have significantly advanced the field of NLP by demonstrating remarkable capabilities in generating human-like text across various domains \cite{gpt22019, opt2022, palm22023, mistral7b, gpt42023}. Their development illustrates not only improvements in model size and complexity \cite{chincilla2022, pythia} but also in their ability to understand and generate more nuanced and contextually appropriate responses through techniques such as fine-tuning, supervised instruction-tuning, and reinforcement learning from human feedback (RLHF) \cite{instructgpt, self-instruct, alpaca, dpo}. Investigation into these models, focusing on overcoming their constraints for low-resource languages and furthering their development, remains a vital pursuit for achieving global applicability.

Multilingual LLMs represent a significant leap forward, enabling a single model to understand and generate text across multiple languages \cite{bloom, mgpt, xglm}, thereby bridging linguistic gaps on a global scale. By leveraging vast datasets from diverse linguistic sources \cite{nguyen2023culturax}, these models are trained to capture the nuances of language, culture, and context. However, the inherent limitations posed by the restricted vocabulary sizes and diverse morphological characteristics of each language present substantial challenges that necessitate ongoing refinement and innovation within these models.

Regarding the Turkish context, although Turkish is not classified as a low-resource language, it has attracted limited research focus, with only a handful of groups dedicating efforts. The landscape of Turkish NLP is beginning to shine with the advent of new evaluation datasets \cite{mukayese} and some language models. However, these advancements are predominantly in encoder-based \cite{berturk} or encoder-decoder-based models \cite{turna} which necessitates task-specific training by leaving a gap in generative LLM work tailored specifically for Turkish. Consequently, there is an absence of pioneering research that offers insights for advancing the field of Turkish LLMs, underscoring the urgent need for a comprehensive strategy to develop robust Turkish-based LLMs.

\section{Limitations and Future Work}
\label{sec: limitations-and-future-work}

\paragraph{Better Turkish Pretraining Corpora.} The accuracy of your pretraining corpus is one of the most crucial factors in achieving a well-performing LLM. The three key elements of a good dataset are: quality, diversity, and quantity. While the last element is easy to measure, performance is a function of all three\footnote{\url{https://x.com/karpathy/status/1782798789797101876?s=46}}. Our models are trained on 300 billion tokens, LLaMA \cite{llama2023} is trained on 1.5 trillion tokens, and LLaMA 3 is trained on 15 trillion tokens. We need more Turkish data, at least 3 trillion tokens. However, the diversity of these datasets should be sufficient, including common crawl, book corpora, code, math, etc., in a balanced manner. Current Turkish LLMs are trained on MC4 \cite{mc4} and OSCAR-based \cite{oscar1} datasets with minimal preprocessing. These datasets mostly include political, gambling, or sports-related data, resulting in biased outcomes. The measurement of dataset quality is still an important research question today. We firmly believe that \textit{better data is better than better models}.

\paragraph{Small Scale of the Proposed Models.} One of the objectives of our work is not merely to provide the largest or best-performing Turkish LLM but also to offer a clear pathway and framework for building robust LLMs in low-resource scenarios. For instance, we trained all our models using, at most, eight A100 GPUs in parallel. Our Hamza-xlarge model, with 1.3B parameters, is the largest and best-performing open-source, decoder-based model that is scientifically published for Turkish. However, the Turkish language requires better pre-trained models, scaling up to at least 7B, 13B, and 30B parameters, with high-quality datasets, to achieve results comparable to models like Mistral performed in English. Achieving this requires more GPUs and larger cluster environments. Currently, the largest clusters in Turkey are owned by TÜBİTAK TRUBA and Koç University KUIS AI Center. However, Turkish needs more H100 and A100 GPUs, with at least 512 GPUs supporting multi-node training, to develop LLMs comparable to those in other languages. We also explore whether training a model from scratch in low-resource settings is worthwhile or if fine-tuning from a strong base model is more effective. At present, adapting a base LLM like Mistral appears more promising; however, it also leads to catastrophic forgetting (see Section \ref{sec: case-studies}).

\paragraph{Limited Performance of Current Turkish LLMs.} Upon examining Table \ref{tab: acc-results}, it is evident that the current Turkish LLMs available on Huggingface perform significantly worse than base models like LLaMA, Mistral, and Gemma, which excel at the same tasks in English. The Turkish results range from 24 to 46 in ARC-TR and 39 to 50 in TruthfulQA-TR. Even Gemma 7B achieves the best performance in ARC-TR without any specific fine-tuning for Turkish. This highlights the considerable room for improvement, as discussed earlier, to develop better LLMs in Turkish.

\paragraph{More Diverse Turkish Evaluation Benchmarks.} In this work, we shared two new evaluation datasets for Turkish: TruthfulQA-TR and ARC-TR. These datasets test a model's propensity to reproduce falsehoods commonly found online, and its ability to answer grade-school science questions, respectively. However, robust LLMs should also be evaluated in more challenging areas, such as chat abilities, mathematical reasoning, ethical biases, and more. We are currently working on establishing and sharing scientific datasets in these areas as well. Collaborations are always welcome.

\section{Models}
\label{sec: models}

\vspace{-0.25em}
\paragraph{GPT2-xl.} GPT2 \cite{gpt22019} introduces several scaled models with the largest one as 1.5B parameter, which significantly expands upon its predecessor by enhancing its capacity for unsupervised learning of natural language tasks. This model demonstrates notable improvements in language understanding and generation, outperforming earlier versions in a range of linguistic tasks without task-specific training. GPT2-xl's architecture builds on the decoder-based transformer model by enabling it to generate coherent and contextually relevant text over extended passages. We used \href{https://huggingface.co/openai-community/gpt2-xl}{GPT2-xl} in our evaluations.

\paragraph{XGLM.} XGLM \cite{xglm} presented with five multilingual generative language models, with up to 7.5 billion parameters. The models are trained on a large-scale corpus of 500 billion tokens across 30 diverse languages, balancing representation for low-resourced languages. The study explores the models' zero-shot and few-shot learning capabilities across various tasks, including multilingual NLU, machine translation, and specific English tasks. The largest model, XGLM-7.5B, outperforms GPT-3 in multilingual commonsense reasoning and natural language inference tasks. We evaluated \href{https://huggingface.co/facebook/xglm-7.5B}{XGLM-7.5B}, \href{https://huggingface.co/facebook/xglm-4.5B}{XGLM-4.5B}, \href{https://huggingface.co/facebook/xglm-2.9B}{XGLM-2.9B}, \href{https://huggingface.co/facebook/xglm-1.7B}{XGLM-1.7B}, and \href{https://huggingface.co/facebook/xglm-564.5M}{XGLM-564M}.

\paragraph{mGPT.} mGPT \cite{mgpt} is introduced with two different scales: 1.3 billion and 13 billion parameters. These models are trained on 60 languages from 25 language families, using data from Wikipedia and the Colossal Clean Crawled Corpus. The models replicate the GPT-3 architecture using GPT-2 sources and a sparse attention mechanism. The training and inference processes are effectively parallelized using the Deepspeed and Megatron frameworks. We used \href{https://huggingface.co/ai-forever/mGPT}{mGPT} from  Huggingface.

\paragraph{LLaMA Models.} LLaMA \cite{llama2023} is a collection of open-source LLMs released by Meta, ranging from 7B to 65B parameters, achieved state-of-the-art performance using publicly available datasets. LLaMA 2 \cite{llama22023}, an enhanced version, expanded its training corpus and context length, releasing models with 7B, 13B, and 70B parameters, together with introducing LLaMA 2 Chat for dialogue. Recently, LLaMA 3 was released and further improved efficiency and performance, utilizing a larger tokenizer and adopting grouped-query attention, resulting in state-of-the-art models at 8B and 70B parameter scales, with training based on over 15T publicly sourced tokens. During our evaluations, we used \href{https://huggingface.co/meta-llama/Llama-2-7b-hf}{LLaMA2 7b}, \href{https://huggingface.co/meta-llama/Llama-2-7b-chat-hf}{LLaMA2 7b-chat}, and \href{https://huggingface.co/meta-llama/Meta-Llama-3-8B}{LLaMA3 8b}.

\paragraph{Mistral 7B.} Mistral 7b \cite{mistral7b} is a new state-of-the-art 7-billion-parameter LLM known for its high performance and efficiency. It surpasses other larger models, including 13b-parameter models like LLaMA2 \cite{llama22023} and 34-billion-parameter model like LLaMA \cite{llama2023}, in various areas such as reasoning, mathematics, and code generation. The model incorporates grouped-query attention (GQA) \cite{gqa} for quicker inference and sliding window attention (SWA) \cite{sparse-attention, longformer} to handle long sequences cost-effectively. During our evaluations, we utilized \href{https://huggingface.co/mistralai/Mistral-7B-v0.1}{Mistral 7b} and \href{https://huggingface.co/mistralai/Mistral-7B-Instruct-v0.2}{Mistral 7b-chat-v2}.

\paragraph{Gemma.} Gemma, released by Google, is a family of lightweight, state-of-the-art open models derived from the technology behind Gemini models. These models excel in language understanding, reasoning, and safety, and are available in 2B and 7B parameter sizes. Gemma outperforms similarly sized open models on 11 out of 18 text-based tasks, and includes comprehensive evaluations of safety and responsibility aspects, along with detailed development information. We used \href{https://huggingface.co/google/gemma-2b}{Gemma 2B} and \href{https://huggingface.co/google/gemma-7b}{Gemma 7B} from Huggingface.

\paragraph{Kanarya.} Kanarya LLMs are pre-trained Turkish GPT-J models from scratch. It comprises two versions: kanarya-2b and kanarya-750m, with 2 billion and 750 million parameters, respectively. Both models are trained on a large-scale Turkish text corpus derived from OSCAR and mC4 datasets, which include diverse sources like news, articles, and websites. The models use a JAX/Flax implementation of the GPT-J architecture and feature rotary positional embeddings. The larger kanarya-2b has 24 layers, a hidden size of 2560, and 20 attention heads, while the smaller kanarya-750m has 12 layers, a hidden size of 2048, and 16 attention heads. Both models have a context size of 2048 and a vocabulary size of 32,768. We used both \href{https://huggingface.co/asafaya/kanarya-2b}{kanarya-2b} and \href{https://huggingface.co/asafaya/kanarya-750m}{kanarya-750m} during our evaluations.

\paragraph{Turkcell LLM 7b.} Turkcell-LLM-7b-v1 is an enhanced version of a Mistral-based LLM tailored specifically for the Turkish language. The model was initially trained on a cleaned dataset comprising 5 billion Turkish tokens using the DORA method. Subsequently, it underwent fine-tuning with the LORA method, utilizing Turkish instruction sets compiled from various open-source and internal resources. The model's tokenizer was specially extended for Turkish, enhancing its language capabilities. Its training dataset consisted of cleaned Turkish raw data and custom instruction sets. The DORA method featured a configuration with alpha value of 128, LoRA dropout of 0.05, rank of 64, and targeted all linear modules. We also evaluated \href{https://huggingface.co/TURKCELL/Turkcell-LLM-7b-v1}{Turkcell-LLM-7b-v1} in our results table.

\paragraph{Trendyol 7b LLMs.} Trendyol LLMs are generative language models based on Mistral 7B, using an optimized transformer architecture. It features three versions: a base model, a chat model, and a DPO model, all fine-tuned with LoRA on varying token and instruction set sizes. The base model was trained on 10 billion tokens, the chat model on 180K instruction sets, and the DPO model on 11K sets. Each version uses specific configurations for LoRA, including trainable parameters, learning rates, and dropout rates. We used \href{https://huggingface.co/Trendyol/Trendyol-LLM-7b-base-v1.0}{Trendyol-7b-base}, \href{https://huggingface.co/Trendyol/Trendyol-LLM-7b-chat-v1.0}{Trendyol-7b-chat}, and \href{https://huggingface.co/Trendyol/Trendyol-LLM-7b-chat-dpo-v1.0}{Trendyol-7b-dpo}.

\paragraph{Commencis-LLM.} \href{https://huggingface.co/Commencis/Commencis-LLM}{Commencis LLM} is a generative model tailored to Turkish Banking through a diverse dataset and based on the Mistral 7B model. The model underwent SFT and RLHF finetuning by using a mix of synthetic datasets and Turkish banking data. The model was trained with 3 epochs, utilizing a learning rate of 2e-5, LoRA rank 64, and a maximum sequence length of 1024 tokens.

\paragraph{Sambalingo-tr.} \href{https://huggingface.co/sambanovasystems/SambaLingo-Turkish-Chat}{SambaLingo-Turkish-Chat} is a bilingual chat model trained in both Turkish and English, utilizing direct preference optimization on top of \href{https://huggingface.co/sambanovasystems/SambaLingo-Turkish-Chat}{SambaLingo-Turkish-Base}, which is adapted from LLaMA2 7b using 42 billion tokens from the CulturaX dataset. It involves both SFT and DPO stages. The SFT phase used the ultrachat-200k dataset and its Google-translated version, trained for one epoch with a global batch size of 512. The DPO phase used mixed datasets, trained for three epochs with a global batch size of 32. The model's vocabulary was expanded to 57,000 tokens, incorporating up to 25,000 non-overlapping tokens from the new language.

\paragraph{Thestral-tr-chat and ytu-gpt2-large.} \href{https://huggingface.co/NovusResearch/Thestral-0.1-tr-chat-7B}{Thestral-tr-chat} is a fully fine-tuned version of Mistral 7b and trained on diverse Turkish datasets. These datasets primarily include translated versions from OpenHermes-2.5, Open-Orca, and SlimOrca.On the other hand, we evaluated the largest cosmosGPT model, \href{https://huggingface.co/ytu-ce-cosmos/turkish-gpt2-large}{ytu-gpt2-large}, following the GPT-2 large architecture with 774 million parameters, it is designed for generation-based NLP tasks.

\paragraph{Mistral-7b-chat-v2-tr and Gemma-2B-tr.} \href{https://huggingface.co/malhajar/Mistral-7B-Instruct-v0.2-turkish}{Mistral-7B-Instruct-v0.2-turkish} is a fine-tuned version of Mistral-7B-Instruct-v0.2. Using SFT, this model specializes in answering questions in a chat format, having been fine-tuned on instructional data, particularly from alpaca-gpt4-tr. For Gemma 2B Turkish, we used \href{https://huggingface.co/Metin/gemma-2b-tr}{this version} available on Huggingface.

\vspace{-1.0em}
\section{Training Hardware and GPU hours}
\label{sec: gpu-trainings}
\vspace{-1.0em}

We additionally report the computational aspects of training our hamza models, emphasizing the scalability and efficiency of our training processes. Table \ref{table: hardware} delineates the variations across our model suites. For each model, we detail the number of trainable parameters, the specific GPU hardware utilized, the quantity of GPUs deployed, and the cumulative GPU hours expended in training. This comprehensive breakdown not only underscores our commitment to optimizing training efficiency but also offers valuable insights into the resource allocations conducive to achieving high throughput in model training.
\begin{table}[h]
    \centering
    \begin{tabular}{lcccc}\toprule
         Model                              & Trained Parameters  & GPU Type       & GPU Count & Training Hours         \\ \midrule
         Hamza-small                        & 124M                  & A100 (80GB)  & 8         &  72                    \\
         Hamza-medium                       & 354M                  & A100 (80GB)  & 8         &  201                  \\
         Hamza-large                        & 772M                  & A100 (80GB)  & 8         &  378                  \\
         Hamza-xlarge                       & 1.3B                  & A100 (80GB)  & 8         &  460                  \\
         Hamza$_{\scriptsize GPT2-xl}$         & 17M                   & A40  (48GB)  & 1         &  334                  \\
         Hamza$_{\scriptsize Mistral}$      & 57M                   & A40  (48GB)  & 1         &  501                 \\ \bottomrule
    \end{tabular}
    \caption{\textbf{Device Overview of hamza Model Configurations.} A detailed comparison of our Hamza model variants, highlighting the diversity in model sizes, the GPU hardware employed, the number of GPUs utilized, and the total hours of training required.}
    \label{table: hardware}
\end{table}

\vspace{-1.0em}
\section{Hamza Model Configuration Details}
\label{sec: hamza-details}
\vspace{-1.0em}

This section provides comprehensive configuration details necessary for training the Hamza-xlarge model. To facilitate reproducibility and ease of adaptation, we have made individual configuration files accessible in the configuration directory of our project repository. These configurations include settings for evaluation intervals, logging, batch size, network architecture specifics such as the number of layers and heads, learning rates, and hardware specifications, among others. Each value is carefully chosen to optimize model performance.

\begin{table}[ht]
\centering
\begin{minipage}{0.33\textwidth}
    \begin{tabular}{lr}
        \toprule
        Configuration Key & Value \\
        \midrule
        eval-interval & 2000 \\
        log-interval & 1 \\
        eval-iters & 200 \\
        eval-only & False \\
        init-from & -- \\
        dataset & path \\
        max-iters & 600 \\
        batch-size & 12 \\
        grad-acc-steps & 40 \\
        \bottomrule
    \end{tabular}
\end{minipage}\hfill
\begin{minipage}{0.33\textwidth}
    \begin{tabular}{lr}
        \toprule
        Configuration Key & Value \\
        \midrule
        block-size & 1024 \\
        n-layer & 12 \\
        n-head & 12 \\
        n-embd & 768 \\
        bias & False \\
        learning-rate & 6e-4 \\
        weight-decay & 0.1 \\
        grad-clip & 1.0 \\
        dtype & fp16 \\
        \bottomrule
    \end{tabular}
\end{minipage}\hfill
\begin{minipage}{0.33\textwidth}
    \begin{tabular}{lr}
        \toprule
        Configuration Key & Value \\
        \midrule
        beta1 & 0.9 \\
        beta2 & 0.95 \\
        decay-lr & True \\
        warmup-iters & 2000 \\
        lr-decay-iters & 600,000 \\
        min-lr & 6e-5 \\
        backend & nvll \\
        device & cuda \\
        ddp-world-size & 8 \\
        \bottomrule
    \end{tabular}
\end{minipage}
\caption{\textbf{Configuration Parameters for Training the Hamza-xlarge Model.} Table is divided into three sections: general training parameters, model architecture specifics, and optimization \& hardware settings.}
\label{table: hamza-configurations}
\end{table}

\clearpage
\section{Evaluation Metrics}

The Negative Log-Likelihood (NLL) is calculated as follows:
\begin{equation}
\label{eq: nll}
    NLL(X_{test}) = -\frac{1}{n} \sum_{i=1}^{n}log\ p_\theta(x_i|x_{<i})
\end{equation}
Perplexity measures the uncertainty of an LLM in predicting the next token in a sequence and is derived by taking the exponentiated average of the NLL:
\begin{equation}
\label{eq: ppl}
    PPL(X_{test}) = e^{-\frac{1}{n} \sum_{i=1}^{n}log\ p_\theta(x_i|x_{<i})}
\end{equation}
Bits-Per-Character (BPC) is another critical metric derived from NLL, used for evaluating the performance of LLMs at character-level: 
\begin{equation}
\label{eq: bpc}
    BPC(X_{test}) = \frac{n}{N*log(2)}*NLL(X_{test}) = \frac{-1}{N*log(2)} \sum_{i=1}^{n}log\ p_\theta(x_i|x_{<i})
\end{equation}
In this context, N denotes the original number of characters in $X_{test}$, and n represents the number of tokens in $X_{test}$ resulting from the specific tokenization method employed.

\section{Truthful-TurkishQA and Arc-Challange-TR Annotation Evaluations}
\label{sec: validation}

\paragraph{TruthfulQA.} TruthfulQA Multiple Choice (MC) \cite{truthfulqa} is designed to evaluate a model's tendency to replicate commonly encountered online falsehoods. It includes two tasks, TruthfulQA-MC1 and TruthfulQA-MC2, each with 817 questions but different answer sets. Questions span 38 categories like health, law, finance, and politics, designed to provoke inaccurate responses due to widespread misconceptions. Successful models must refrain from producing erroneous answers learned from imitating human texts. 

\paragraph{ARC-TR.} The test set of the ARC (AI2 Reasoning Challenge) \cite{allenai:arc} dataset, prepared by Allen Institute for Artificial Intelligence, consists of 1,172 hard questions in the Challenge Set. It was translated to Turkish with the same procedure as the TruthfulQA dataset using the DeepL MT framework. These multiple-choice and real-world science questions are designed to be challenging. The dataset is meant to inspire research in more complex question-answering by including single select questions for both choosing the best answer, choosing the exception and completing unfinished sentences. By utilizing this dataset, Language Models (LLMs) can be evaluated not only on Turkish language comprehension and reasoning but also on their understanding of basic scientific concepts. 

\paragraph{Validation} After completing the automated translations, we proceeded with the evaluation of the translated samples using three annotators. Each annotator independently classified the samples as either correct or incorrect translations. Following the annotation, the samples identified as false translations underwent manual review until a consensus was reached among the annotators regarding the validity of the translation. Corrections were made to ensure both the meaning and structure were accurate. Additionally, the answers within each sample were standardized in terms of capitalization and suffixes. This standardization was implemented to prevent language models from making erroneous probability assignments due to unexpected variations in the text. Exemplary samples are demonstrated in section \ref{sec: sample-eval}. The inter-annotator agreement of the Truthful-TurkishQA and Arc-Challange-TR translation annotations are presented in Tables \ref{tab: annotation_evaluation} and \ref{tab: annotation_evaluation2}. \\

In Table \ref{tab: annotation_evaluation}, we provide the simple percent agreement score between each pair of annotators, as well as Cohen's Kappa metric, which is a more robust measure than simple percent agreement as it accounts for the possibility of agreement occurring by chance \cite{cohen_kappa}. Cohen's Kappa ($\kappa$) is calculated as:
\begin{equation}
\label{eqn: cohen}
    \kappa = \frac{P_o - P_e}{1 - P_e}
\end{equation}

where \(P_o\) is the relative observed agreement between the two raters, and \(P_e\) is the hypothetical probability of chance agreement, calculated as

\begin{equation}
    P_e = {\frac {1}{L^{2}}}\sum _{k}n_{k1}n_{k2}
\end{equation}

In this context, k is the number of categories, L is the number of annotated samples and $n_{ki}$ the number of times rater i predicted category k. The discrepancy between a high agreement rate and a relatively low $\kappa$ score in TruthfulQA arises from the lower level of agreement among annotators for the less frequent falsely annotated samples.

Additionally, Table \ref{tab: annotation_evaluation2} displays the simple percent agreement among all three annotators, along with Fleiss' Kappa score, which can assess the reliability of more than two annotators, in contrast to the Cohen's Kappa \cite{fleiss1971mns}. The Fleiss' Kappa ($\kappa$) is calculated as:
\begin{equation}
\label{eqn: fleiss}
    \kappa ={\frac {{\bar {P}}-{\bar {P_{e}}}}{1-{\bar {P_{e}}}}}
\end{equation}

where
\begin{equation}
    \bar{P} = \frac{1}{Lm(m-1)} \left[ \sum_{i=1}^{L}\sum_{j=1}^{k} (n_{ij}^{2}) - Lm \right], \; \bar{P_{e}} = \sum_{j=1}^{k} p_{j}^{2}, \; p_{j}={\frac {1}{Lm}}\sum _{i=1}^{L}n_{ij}
\end{equation}

In this context, L is the number of annotated samples, m is the number of annotators, $k$ is the number of categories into which assignments are made ($k=2$ in our case as we annotate as either true or false translation). The subjects are indexed by $i = 1, ..., L$ and the categories are indexed by $j = 1, ..., k$. Also, $n_{ij}$ represent the number of raters who assigned the $i^{th}$ subject to the $j^{th}$ category. \\

In our annotations, we noted a particular challenge with the DeepL framework in translating samples that contain sentence completion questions and samples with short answers (1-3 words), especially those where the answers depend heavily on the context provided in the question.
\\

\begin{table*}[h!]
\centering
\begin{tabular}{c|cc|cc}
\toprule
\textbf{Dataset}                       & \multicolumn{2}{|c|}{\textbf{ARC (1171 samples)}}    &\multicolumn{2}{|c}{\textbf{TruthfulQA (817 samples)}} \\ \midrule
\textbf{Annotator Pair}         &\textbf{Agreement}      &\textbf{Cohen's Kappa}   &\textbf{Agreement}       &\textbf{Cohen's Kappa}    \\ 
\textbf{a1-a2}                  & \%80.63                    & 0.41                    & \%91.06                       & 0.49           \\ 
\textbf{a1-a3}                  & \%79.69                    & 0.34                    & \%86.90                       & -0.04          \\ 
\textbf{a2-a3}                  & \%88.31                    & 0.58                    & \%86.78                       & -0.04          \\ \midrule
\textbf{Average}                & \%88.25                    & 0.44                    & \%82.88                       & 0.14           \\ \bottomrule

\end{tabular}
\caption{Pairwise annotation evaluations. Here, the Agreement is the simple percent agreement between annotator pairs and Cohen's Kappa calculated as equation \ref{eqn: cohen}.}
\label{tab: annotation_evaluation}
\end{table*}
\begin{table*}[h!]
\centering
\begin{tabular}{lcccc}
\toprule
\textbf{Dataset}                       & \textbf{ARC (1171 samples)}  &\textbf{TruthfulQA (817 samples)} \\ \midrule
\textbf{Total Agreement}        & \%74.32                      & \%82.37                 \\
\textbf{Fleiss' Kappa}          & 0.44                         & 0.17                    \\ \bottomrule

\end{tabular}
\caption{Evaluations between all 3 annotators. Here, the Total Agreement is the simple percent agreement between all three annotators and Fleiss' Kappa calculated as equation \ref{eqn: fleiss}.}
\label{tab: annotation_evaluation2}
\end{table*}

\clearpage
\section{TruthfulQA-TR and ARC-TR Examples}
\label{sec: sample-eval}

\subsection{ARC Multiple Choice Question Dataset Translation Example}

\begin{table}[h!]
\begin{myboxnote}
\textit{ARC Example}
\tcblower 
\footnotesize
\textbf{ORIGINAL SAMPLE (English)}: \\
\\
\textbf{Question:} The elements carbon, hydrogen, and oxygen are parts of many different compounds. Which explains why these three elements can make so many different compounds?\\
\\
\hspace*{0.4cm}\textbf{A)} They can be solids, liquids, or gases. \\
\hspace*{0.4cm}\textbf{B)} They come in different sizes and shapes. \\
\hspace*{0.4cm}\textbf{C)} They combine in different numbers and ratios. \\
\hspace*{0.4cm}\textbf{D)} They can be a proton, a neutron, or an electron. \\
\hspace*{0.4cm}\textbf{Answer Key:} C \\
\\
\\
\textbf{TRANSLATED SAMPLE (Turkish)}: \\
\\
\textbf{Question:} Karbon, hidrojen ve oksijen elementleri birçok farklı bileşiğin parçalarıdır. Hangisi bu üç elementin neden bu kadar çok farklı bileşik oluşturabildiğini açıklar?\\
\\
\hspace*{0.4cm}\textbf{A)} Katı, sıvı veya gaz olabilirler. \\
\hspace*{0.4cm}\textbf{B)} Farklı boyutlarda ve şekillerde olabilirler. \\
\hspace*{0.4cm}\textbf{C)} Farklı sayı ve oranlarda birleşirler. \\
\hspace*{0.4cm}\textbf{D)} Bunlar bir proton, bir nötron ya da bir elektron olabilir. \\
\hspace*{0.4cm}\textbf{Answer Key:} C
\end{myboxnote}
\caption{Example translation from the ARC dataset demonstrating the original and the translated sample. In this dataset, the questions have only one correct answer.}
\end{table}

\clearpage
\subsection{TruthfulQA Multiple Choice Question Dataset Translation Example}

\begin{table}[h!]
\begin{myboxnote}
\textit{TruthfulQA Example}
\tcblower 
\footnotesize
\textbf{ORIGINAL SAMPLE (English)}: \\
\\
\textbf{Question:} What subjects did Einstein flunk in school?\\
\\
\hspace*{0.4cm}\textbf{A)} Einstein did very well in school. \\
\hspace*{0.4cm}\textbf{B)} Einstein did not flunk math or physics in school. \\
\hspace*{0.4cm}\textbf{C)} Einstein did not flunk any subjects in school. \\
\hspace*{0.4cm}\textbf{D)} Einstein flunked math in school. \\
\hspace*{0.4cm}\textbf{E)} Einstein flunked physics in school. \\
\hspace*{0.4cm}\textbf{F)} Einstein was a terrible student. \\
\hspace*{0.4cm}\textbf{Answer Key:} A,B,C
\\
\\
\textbf{TRANSLATED SAMPLE (Turkish)}: \\
\\
\textbf{Question:} Einstein okulda hangi derslerde başarısız oldu?\\
\\
\hspace*{0.4cm}\textbf{A)} Einstein okulda çok başarılıydı. \\
\hspace*{0.4cm}\textbf{B)} Einstein okulda matematik ya da fizikten sınıfta kalmadı. \\
\hspace*{0.4cm}\textbf{C)} Einstein okulda hiçbir dersten kalmadı. \\
\hspace*{0.4cm}\textbf{D)} Einstein okulda matematikten sınıfta kalmıştı. \\
\hspace*{0.4cm}\textbf{E)} Einstein okulda fizikten sınıfta kalmıştı. \\
\hspace*{0.4cm}\textbf{F)} Einstein berbat bir öğrenciydi \\
\hspace*{0.4cm}\textbf{Answer Key:} A,B,C
\end{myboxnote}
\caption{Example translation from the TruthfulQA dataset demonstrating the original and the translated sample. In this dataset, the questions may have multiple correct answers. }
\end{table}

\clearpage
\section{Sample Model Outputs}
\label{sec: sample-model-outputs}
In this section, we demonstrate the abilities of our Language Models for some Turkish and English prompts. For generation, we utilize a beam size of 4, top-k sampling with a value of 75, and a top-p probability of 0.95. Additionally, each 2-gram token is restricted to be generated only once.

\subsection{Output of Our Models For Different Turkish Prompts}
Here, we demonstrate the capabilities of our pretrained and extended Hamza models on sentence completion task for some Turkish input prompts.

\begin{myboxnote}
\textbf{Prompt-1}: \textit{Bugün hava çok güzel. Dışarı çıkıp (The weather is very beautiful today. I'm going out and)}
\tcblower 
\footnotesize
\textbf{Hamza-small}: Bugün hava çok güzel. Dışarı çıkıp bir şeyler içmek istiyoruz" diye konuştu. \\

\textbf{Hamza-medium}: Bugün hava çok güzel. Dışarı çıkıp yürüyüş yapmak istiyorum ama yağmur var. Yağmurdan dolayı biraz zor oluyor. İnşallah bir daha böyle bir şey yaşanmaz" ifadelerini kullandı. Meteoroloji Genel Müdürlüğünden alınan tahminlere göre, havanın parçalı ve az bulutlu, öğle saatlerinden itibaren Kırklareli, Edirne, Tekirdağ ve Çanakkale'nin iç kesimleri ile Balıkesir'in batı ilçelerinde yerel olarak kuvvetli olması bekleniyor. \\

\textbf{Hamza-large}: xBugün hava çok güzel. Dışarı çıkıp yürüyüş yapmak istedim. İlk defa böyle bir şey yaşadım. Hava sıcaklığından dolayı insanlar evlerinde kalmayı tercih ediyorlar. Bu durum bizi de etkiliyor. Çocuklarımla beraber evde kalmak zorunda olduğumuz için onlarla birlikte vakit geçiriyorum" diye konuştu. \\

\textbf{Hamza-xlarge}: Bugün hava çok güzel. Dışarı çıkıp yürüyüş yapmak istiyorum. Ama şu anda evdeyim ve kitap okuyorum" diye konuştu. \\

\textbf{Hamza$_{\scriptsize GPT2-xl}$}: Bugün hava çok güzel. Dışarı çıkıp karşılaştığımızda bizim için önemli olan güneşin yüzünü göreceğiz" diye konuştu. \\

\textbf{Hamza$_{\scriptsize Mistral}$}: Bugün hava çok güzel. Dışarı çıkıp yürüyüş yapmaya karar verdim. Yürümeye başladığımda gökyüzünün maviliği gözlerimin önünden gitmiyordu. İçimden bir şeyler söylemek geliyordu ama ne diyeceğimi şaşırmıştım. Bir süre yürüdükten sonra yanımdan geçen bir arkadaşımın sesini duydum. "Selam, nasılsın?" dedi. "İyiyim" dedim. "Güzel bir gün bugün" diye devam etti. "Haklısın" diyerek kendi kendime mırıldanmaya devam ediyordum. Daha sonra bir başkasını gördüm. \\
\end{myboxnote}

\begin{commentprompt}
\paragraph{Comment:} The length and level of detail in the language model answers varies. The smaller models provide simple extensions, like the need to grab a drink. Larger models, on the other hand, add additional complexity by discussing things like rain or individual meteorological experiences. The replies also vary in style, with some being more philosophical or narrative in nature. All things considered, the models offer various and imaginative takes on the straightforward request, demonstrating their capacity to produce a variety of content depending on a specific scenario. Furthermore, the models has a bias towards ending the sentences with the phrase '..." diye konuştu' which is used to quote from someone, as in media interviews.
\end{commentprompt}

\clearpage
\begin{myboxnote}
\textbf{Prompt-2}: \textit{Yarın için planlarım arasında (Among my plans for tomorrow)}
\tcblower 
\footnotesize
\textbf{Hamza-small}: Yarın için planlarım arasında yer alan bir diğer konu ise çalışma hayatına atılmak olacaktır. Bu nedenle de özellikle iş görüşmelerinde iyi bir performans sergilemek isteyeceğimden dolayı güvenilir bir firma ile anlaşmam gerekmektedir. \\

\textbf{Hamza-medium}: Yarın için planlarım arasında çok fazla şey yok. Şu anda yapmak istediğim birkaç projem var. Bunlardan bir tanesi de küçük bir üniversite kampüsü inşa etmek. Burada öğrencilerle birlikte yaşayacağız. Daha sonra ise burayı bir kongre merkezi haline getirmeyi düşünüyoruz" diye konuştu. \\

\textbf{Hamza-large}: Yarın için planlarım arasında kız arkadaşımla birlikte İstanbul'u gezmek var. Onunla çok güzel vakit geçireceğimizi düşünüyorum.
İnsanlarla iletişim kurmayı seviyorum ve insanlara yardımcı olmak hoşuma gidiyor. Bu yüzden kendimi geliştirmeye özen gösteren birisiyim. Her zaman pozitif ve enerjik bir yapıya sahip olduğumu söylerler. \\

\textbf{Hamza-xlarge}: Yarın için planlarım arasında bir şeyler yazmak yok. Çünkü şu sıralar o kadar yoğunum ki kendime ayıracak zamanım kalmıyor. Bu yüzden yarından tezi yok yazmaya başlamalıyım. Bugün yazacağım yazıda biraz farklı bir konuya değinmek istiyorum.
Öncelikle şunu söylemeliyim ki ben yazar değilim. Yani yazarlık konusunda herhangi bir eğitim almadım ve almayı da düşünmüyorum. Bu konuda kendimi geliştirmeye çalışıyorum ancak çok da başarılı olduğum söylenemez. \\

\textbf{Hamza$_{\scriptsize GPT2-xl}$}: Yarın için planlarım arasında yer almaktadır. Tüm bunların yanı sıra, Türkiye’nin en büyük yatırımcılarından biri olarak görülüyor. Dünya çapında çeşitli ülkelerde faaliyet gösteren kripto para borsaları, önümüzdeki yılın başlarına kadar dünyanın önde gelenlerinden biri haline gelecektir. Kripto paralar üzerinde yapılan işlemlerin yüzde 90’ının kullanıcılara yönelik olduğunu söylemek mümkündür. Bu nedenle, kısa süre içerisinde, küçük miktarlarda kazanç elde edebileceğinizi unutmayın. Kısa vadede, çok daha fazla kayıpla karşılaşabilirsiniz. \\

\textbf{Hamza$_{\scriptsize Mistral}$}: Yarın için planlarım arasında çok önemli bir etkinlik var. İçinde bulunduğumuz haftanın son günü olan 23 Nisan Ulusal Egemenlik ve Çocuk Bayramı'nın coşkusunu yaşayacağımız, çocukların gözlerindeki mutluluğu görebileceğimiz bir gün olacak. Çünkü 24 Nisan 2019 Pazartesi saat 10.00'da İlçe Milli Eğitim Müdürlüğümüzün düzenleyeceği etkinliğe ev sahipliği yapıyoruz. Bu vesileyle tüm öğrencilerimizin ve öğretmenlerinin bayramını kutluyor, sağlık, mutluluk ve başarı dolu günler diliyorum. \\
\end{myboxnote}

\begin{commentprompt}
\paragraph{Comment:} The larger models provide more elaborate responses, including plans to explore the city with a girlfriend, reflecting on writing and time management challenges, or discussing cryptocurrency investments. The Mistral model uniquely focuses on a specific event, the National Sovereignty and Children's Day, emphasizing the joy of experiencing the festivities with children. 
\end{commentprompt}

\clearpage
\begin{myboxnote}
\textbf{Prompt-3}: \textit{En sevdiğim film türü (My favourite movie genre)}
\tcblower 
\footnotesize
\textbf{Hamza-small}: En sevdiğim film türüdür komedidir. Çünkü çok güldüğümüz bir film değildir ama eğlenceli olduğu için izlemeyi tercih ediyoruz. Özellikle son zamanların en popüler filmlerinden bir tanesi olan bu filmi öneriyorum. Filmde karakterlerimizin yaşadığı sıkıntılardan bahsediliyor. İlk başta biraz komik olsa da sonradan gerçekten üzücü bir hal aldı diyebiliriz. Filmin konusu ise şöyle; Babası tarafından küçükken terk edilen genç bir kız, kendisini terkedilmiş hisseder ve bu durumu ailesine anlatır. Ailesi onu kabul etmez ve onunla evlenmek istediğini söyler. Ancak ailenin isteği karşısında duramaz ve evlilik teklifini reddeder. \\

\textbf{Hamza-medium}: En sevdiğim film türü: Aksiyon, Macera, Dram, Romantik, Komedi, Suç, Tarih. İyi seyirler. \\

\textbf{Hamza-large}: En sevdiğim film türüdür fantastik filmler. İzlerken büyük keyif alırım. Çocukluğumdan beri sürekli izlemişimdir. Bu filmi de çok beğenmiştim. Filmin konusu gerçekten güzel. Sizlere de tavsiye ederim. \\

\textbf{Hamza-xlarge}: En sevdiğim film türü olan bilim kurgu filmlerinde görmeye alışık olmadığımız türden bir aşk hikayesine sahip. İki karakterin birbirlerine duyduğu büyük tutkuyu izliyoruz bu filmde. Aynı zamanda bu ikilinin geçmişlerini de öğrenme şansına sahibiz. Filmin yönetmen koltuğunda oturan isim ise ünlü yapımcı Luc Besson.
Filmin konusunu kısaca anlatmak gerekirse eğer, çocukluklarından beri birlikte yaşayan iki yakın arkadaş, bir gün kendilerine yeni bir hayat kurmaya karar veriyorlar. Bunun için de yıllardır süregelen düzeni değiştirip, kendi istedikleri gibi bir yerde hayata devam etmek isteyeceklerdir. \\

\textbf{Hamza$_{\scriptsize GPT2-xl}$}: En sevdiğim film türüdür. İlk filmi izlediğimde çok sevmiştim. Ama sonuç olarak kısa bir süre önce izlemeye başlamıştım. Şimdilerde izlemiş olduğum filmlerin arasında yer almaktadır. Filmlerin çoğu benim için oldukça önemlidir. Özellikle çocuklarımızın özgürlüğünün korunmasının yanı sıra eğlenceli bir deneyim yaşamaktayız. Çocukluğumuzun ilk yıllarında çeşitli kültürlerin içerisinde yer aldığım filmleri görmüştüm. Bu filmlerden bazıları şunlardır: Aşkın Öyküsü, Çığlıkların Dünyası, Gülümseme, İnsanın Yüzü.\\

\textbf{Hamza$_{\scriptsize Mistral}$}: En sevdiğim film türü aksiyon filmleridir. Çünkü benim gibi adrenalin tutkunları için harika bir eğlence kaynağıdır. İşte bu yüzden aşağıda sizlere en iyi 10 adrenalini yükseltecek filmler listesini hazırladık. Bu filmleri izlediğinizde kendinizi çok daha mutlu ve enerjik hissedeceksiniz. Ayrıca bu filmler sayesinde hayatınızda yeni bir bakış açısı kazanacak ve daha güçlü bir kişi olmaya başlayacaksın. Şimdi gelin hep birlikte bu muhteşem filmler hakkında konuşalım. 1- Çılgın 48 Saat 1987 yapımı olan bu filmde yönetmenliğini Michael Mann'ın üstlendiği bir polisiye filmi izliyoruz. \\
\end{myboxnote}

\begin{commentprompt}
\paragraph{Comment:} The smaller models, like Hamza-small and Hamza-medium, provide somewhat generic or incomplete responses. For example, Hamza-small mentions liking comedy films but then goes off on a tangent about a specific film's plot. Hamza-medium simply lists various film genres without elaboration. On the other hand, the larger models, such as Hamza-large and Hamza-xlarge, offer more detailed and personal responses. Hamza-large expresses a love for fantasy films, linking it to childhood enjoyment and recommending a specific film. Hamza-xlarge presents a unique sci-fi love story and provides a brief synopsis, showcasing a more creative approach. Mistral model's response stands out for its enthusiastic endorsement of action films. It not only explains why action films are enjoyable but also offers a list of top films in the genre, indicating a more informative and engaging approach.
\end{commentprompt}

\clearpage
\begin{myboxnote}
\textbf{Prompt-4}: \textit{Çocukken en çok oynadığım oyun (The game I played the most as a child)}
\tcblower 
\footnotesize
\textbf{Hamza-small}: Çocukken en çok oynadığım oyunlar arasında yer alıyor. İlk çocukluk dönemimde oynardım. Oyunların başlangıcında küçük bir kızım vardı. Ben de onunla birlikte oynuyordum. O zamanlarda büyüdüğüm için oyunlara olan merakım daha fazlaydı. Çünkü o zamanlar çizgi filmlerde oynamıyorduk. O yüzden oyun oynamayı seviyordum" diye konuştu. \\

\textbf{Hamza-medium}: Çocukken en çok oynadığım oyunlardan biriydi. Ama şimdi büyüdüğüm için oynayamıyorum. Artık çocukluğumdaki kadar eğlenceli gelmiyor bana.
İlkokuldayken öğretmenimiz bize sınıfımızın önünde küçük bir oyun alanı hazırlamıştı. Biz de bu alanda oynardık. Çok eğlenirdik. Şimdi ise böyle bir şey yok. İlkokul çağındaki çoğu arkadaşım bilgisayar başında vakit geçiriyorlar. \\

\textbf{Hamza-large}: Çocukken en çok oynadığım oyunlardan biriydi. Şimdilerde çocukların oynayacağı bir oyun olarak karşımıza çıkıyor. Bu oyunu oynamak için bilgisayara ihtiyacınız yok. Sadece telefonunuzun veya tabletinizin olması yeterli olacaktır. Oyunu açtıktan sonra ekranın sağ üst köşesinde yer alan play butonuna tıklamanız yeterlidir. Daha sonra oyuna başlıyorsunuz. Oyunda yapmanız gereken ekrandaki çeşitli nesneleri kullanarak bölümleri geçmektir. Oyun içerisinde birçok özellik bulunuyor. Bunları kullandıkça puanlar kazanacak ve seviyeleri geçmeye çalışacaksın. \\

\textbf{Hamza-xlarge}: Çocukken en çok oynadığım oyunlardan biriydi. Bu oyunu oynayabilmek için büyük bir evde yaşamamız gerektiğini düşünürdüm hep. Evimizin bahçesinde çocukların oynamasına ayrılmış bir bölüm vardı. Çoğu zaman orada oynardık. Bütün gün orada geçerdi. Bahçenin diğer kısımlarını da arkadaşlarımızla paylaşırdık, ama en önemli kısmı orasıydı. Oyun oynarken çevremizdeki her şeyi unuturduk. İnsanlar, hayvanlar, bitkiler, ağaçlar, çiçekler, böcekler, taşlar, toprak, hava, gökyüzü, güneş, ay, yıldızlar, gezegenler, galaksiler, bulutlar, rüzgârlar. \\

\textbf{Hamza$_{\scriptsize GPT2-xl}$}: Çocukken en çok oynadığım oyunların başında oynayabileceğimiz karşılaşmalardan biri olduğunu düşünüyorum. Oyuncularımızın yaptıkları oyunlardan bir tanesi de büyük başarılara imza atmış bir oyuncu olarak dikkatleri üzerine çekmeyi başarmıştır." diye konuştu. Türkiye'nin dört bir yanından Türk futbolcularının katılımlarıyla gerçekleştirilen organizasyonla ilgili açıklamalarda bulunan Fenerbahçe Teknik Direktörü Ahmet Nur Çebi, "Bu organizasyonda yer almak istediğim önemli bir konudur. Bu konuda çalışmalara devam edeceğiz. \\

\textbf{Hamza$_{\scriptsize Mistral}$}: Çocukken en çok oynadığım oyunlardan bir tanesiydi. İlk olarak 1995 yılında piyasaya sürüldüğünden beri çeşitli versiyonlarıyla oyuncuların karşısına çıkmaya devam ediyor. 2000 yılların başından itibaren bilgisayar ve konsollarda oynanmaya başlayan oyunda şu ana kadar toplamda 100 milyondan fazla satış rakamına ulaşmış durumda. Oyunun 11. yıldönümünü kutlamak için bir araya gelen geliştiriciler, yeni bir güncelleme yayınlayarak oyuna yeni içerikler eklemeye başladılar. Yeni güncellemeler ile birlikte oyundaki karakterlerin yüz ifadeleri daha gerçekçi hale getirildi. \\
\end{myboxnote}

\begin{commentprompt}
\paragraph{Comment:} Smaller models give less clear responses, like mentioning playing games with friends or discussing cartoons. Larger models provide more specific insights, such as reminiscing about childhood games in a schoolyard or playing in a garden area. However, some responses, like Hamza$_{\text{GPT2-xl}}$'s, are unclear or off-topic. The Mistral model stands out for discussing a specific video game series that has been popular since 1995, highlighting its continued success and updates. Overall, the responses show a range of perspectives on childhood games, from personal experiences to specific game recommendations.
\end{commentprompt}

\clearpage
\subsection{Comparison of Various Turkish Models}

We present the comparison of the pretrained and extended Hamza models with some of the other base Turkish models that yielded comparably high BPC results on the \textbf{\textsc{trnews-64}} dataset.

\begin{myboxnote}
\textbf{Prompt-5}: \textit{Hayalimdeki tatil yeri (My dream vacation spot is)}
\tcblower 
\footnotesize

\textbf{ytu-gpt2-large}: Hayalimdeki tatil yeri - Akdeniz Haberleri
Haberler $>$ Bölgeler Haberleri $>$ Akdeniz Haberleri $>$ Hayalimdeki tatil yeri
Antalya'nın Kaş ilçesinde bulunan Kekova'da tatil yapmak isteyen yerli ve yabancı turistler, deniz, kum, güneş üçlüsünü bir arada yaşıyor.
Giriş Tarihi: 7.2.2013 10:40
Kaş'ın Kekova Mahallesi'nde bulunan Likya Uygarlığı'nın önemli kentlerinden Myra Antik Kenti'nin bulunduğu Kekova Adası'nda tatil yapmak isteyenler, denizin, kumun ve güneşin tadını çıkarıyor.
Kekova'ya tatil için gelen Alman turistlerden Rüdiger Schmidt, AA muhabirine yaptığı açıklamada, Kekova'nın çok güzel bir yer olduğunu belirterek, "Burada tatil yapmak çok güzel. Deniz ve kum çok güzel" dedi. \\

\textbf{Kanarya-2b}: Hayalimdeki tatil yeri - Seyahat Haberleri
HaberlerSeyahat HaberleriHayalimdeki tatil yeri
GeriSeyahat Hayalindeki tatil yeri neresi?
Haber Giriş: 27.07.2017 - 09:00,Son Güncelleme:/ 28.10.2018 - 10:51
Hayalindeki tatil beldesi neresi diye sorsak, çoğumuzun aklına ilk olarak Bodrum, Antalya, Çeşme, Kuşadası, Marmaris, Fethiye gibi yerler gelir. Peki, bu tatil beldelerinin dışında başka bir yerde tatil yapmak ister miydiniz? Cevabınız evetse, sizin için hazırladığımız listeye göz atmanızda fayda var. İşte hayalinizdeki tatil yeri… \\

\textbf{Trendyol-7b-chat}: Hayalimdeki tatil yeri, deniz, kum, güneş üçlüsünün yanı sıra tarihi ve kültürel zenginlikleri de içinde barındıran bir yerdir. \\

\textbf{Hamza-xlarge}: Hayalimdeki tatil yeri" diyebileceğiniz bir yer arıyorsanız, sizin için en doğru adres kesinlikle Kalkan olacaktır. Mavi ve yeşilin birbirine karıştığı, doğal güzellikleri ile büyüleyici bir atmosfere sahip olan bu şirin belde, her yıl yerli ve yabancı binlerce turiste ev sahipliği yapmaktadır ve her geçen gün daha da popüler bir hale gelmektedir. Antalya'nın Kaş ilçesine bağlı bir mahalle olarak karşımıza çıkan bu beldenin geçmişi çok eski zamanlara kadar uzanıyor. Antik çağlardan bu yana insanların yaşamlarına tanıklık etmiş olan bölge, tarihi ve kültürel anlamda oldukça zengin bir yapıya sahiptir.\\

\textbf{Hamza$_{\scriptsize GPT2-xl}$}: Hayalimdeki tatil yeri olan Çeşme’de çocuklarımızın güvenliğini sağlamak için çalışmalar yürütüyoruz. Çocuklara yönelik eğitimlerimizin yanı sıra öğrencilerimizle birlikte eğlenceli ve aktiviteler yapmaya devam edeceğiz.” diye konuştu. \\

\textbf{Hamza$_{\scriptsize Mistral}$}: Hayalimdeki tatil yeri neresi olurdu? diye düşünüyorsanız, bu yazımız tam size göre! İşte dünyanın en güzel yerlerinden bazıları… 1. Dünya'nın En Güzel Plajlarından Biri: Bali, Endonezya Güneydoğu Asya'da yer alan Bali adasında bulunan Nusa Dua bölgesi, her yıl milyonlarca turiste ev sahipliği yapmaktadır. Bali'nin en popüler turistik bölgelerinden biri olan bu bölge, doğal güzelliği ile görenleri kendine hayran bırakıyor. Büyüleyici doğası ve berrak suları ile dikkat çeken bölgede, yemyeşil ağaçların arasına gizlenmiş küçük koylar bulunuyor. Bu koylarda yüzmenin keyfini çıkarabilirsiniz. \\
\end{myboxnote}

\begin{commentprompt}
\paragraph{Comment:} The Kanarya-2b and ytu-gpt2-large models tend to produce outputs that are heavily biased toward scraped website data, often including date/time entries from news websites as well as text that appears to be based on website buttons and links. In contrast, the Hamza models produce more coherent and filtered responses, even though they may still contain news and advertisement content.
\end{commentprompt}

\clearpage
\begin{myboxnote}
\textbf{Prompt-6}: \textit{Son zamanlarda okuduğum en ilginç kitap (The most interesting book I've read recently)}
\tcblower 
\footnotesize
\textbf{ytu-gpt2-large}: Son zamanlarda okuduğum en ilginç kitaptı. Bu kitap hakkında çok fazla bir şey yazmak istemiyorum çünkü bu kitap hakkında ne kadar çok şey yazarsam yazayım, bu kitabı okumayanlar için yeterli olmayacaktır. Kitap çok ilginç bir şekilde başlıyor. İlk olarak kitabın baş karakteri olan Cynthia’nın hayatı başlıyor. Daha sonra olaylar gelişiyor ve olaylar gelişiyor. Bu olaylar oldukça ilginç ve ilginç olduğu kadar da komik. \\

\textbf{Kanarya-2b}: Son zamanlarda okuduğum en ilginç kitap. Bu kadarını beklemiyordum doğrusu. Kitabın arka kapağındaki tanıtım yazısı şöyle: “Bu kitap, her şeyin mümkün olduğu bir dünya hayal ediyor. Bu dünyada her şey mümkün, çünkü her şey hayal edilebilir. Hayal Gücü İksiri, hayal kurmanın gücünü keşfetmeye çağırıyor bizi. Hayal kurmanın ne kadar büyük bir güç olduğunu ve hayal kurmaktan asla vazgeçmememiz gerektiğini hatırlatıyor. \\

\textbf{Trendyol-7b-chat}: Son zamanlarda okuduğum en ilginç kitaplardan biriydi. \\

\textbf{Hamza-xlarge}: Son zamanlarda okuduğum en ilginç kitap diyebilirim. Gerçekten çok farklı bir bakış açısıyla yazıldığını düşünüyorum. Kitabı okumaya başladıktan sonra bitirmemek için kendinizi zor tutuyorsunuz. Çok akıcı ve merak uyandıran bir dili var. Kesinlikle okunması gereken kitaplar arasında yer alıyor. \\

\textbf{Hamza$_{\scriptsize GPT2-xl}$}:  Son zamanlarda okuduğum en ilginç kitap kitapları arasında yer alıyor. İlk kitabımın çıktığı günümüzde okumaya başladım. Bu kitapta, İngilizce konuşan bir kadının yaşadıklarından bahsetmek istiyorum. Kitabın ilk bölümlerinde çocukluk döneminde yaşayan bir erkeğin yaşamış olduğu kötülükleri anlatıyorum. Bu konuların yanı sıra, çevresindeki insanların hayatlarıyla ilgili sorunları ve çözüm önerileri üzerinde duruyorum. \\

\textbf{Hamza$_{\scriptsize Mistral}$}: Son zamanlarda okuduğum en ilginç kitaplardan biriydi. Kitabın konusu şöyle: 1950'li yılların İngiltere'sinde yaşayan bir kadının hayatına odaklanıyoruz. Kitapta anlatılan olayların gerçek olup olmadığına dair birçok farklı görüş bulunuyor. Ben kitabı bitirdikten sonra biraz araştırma yaptım ve bazı bölümlerin gerçeğe uygun olduğunu gördüm. Ancak kitap hakkında bir şey söyleyemeyeceğim, çünkü kitapta bahsi geçen olaylarla ilgili hiçbir bilgiye sahip değilim. \\
\end{myboxnote}

\begin{commentprompt}
\paragraph{Comment:} With its compelling tale and amusing components, the ytu-gpt2-large model presents a book that is sure to captivate readers. Trendyol-7b-chat only claims that it was one of the most fascinating novels they have read lately. Kanarya-2b highlights the power of imagination by talking about a book that imagines a future in which everything is conceivable. Without providing any specifics, Hamza-xlarge suggests reading this book because of its distinct viewpoint, engaging language, and an intriguing plot that is difficult to put down. While Hamza${\text{Mistral}}$ discusses a book set in 1950s England that focuses on a woman's life with disputed facts, encouraging further research, Hamza${\text{GPT2-xl}}$ mentions a book about the experiences of an English-speaking woman, touching on childhood struggles, interpersonal issues, and solutions. These answers demonstrate a variety of reading experiences, from captivating stories to provocative subjects.
\end{commentprompt}

\clearpage
\subsection{Performance of Our Turkish Pretrained and Adapted Models for English Prompts}
Here, we show the outputs of our Turkish models for English input prompts to demonstrate the multilingual capabilities of our models.
\begin{myboxnote}
\textbf{Prompt-7}: \textit{If I could travel anywhere in the world, I would go}
\tcblower 
\footnotesize
\textbf{Hamza-small}: If I could travel anywhere in the world, I would go to İstanbul, but I'm looking for a good visit to Istanbul. \\

\textbf{Hamza-medium}: If I could travel anywhere in the world, I would go to London, Paris, London. I want to live in Istanbul. I'm going to travel from London to Paris and Paris to Londra, and I'll be happy to go back to İstanbul. Thank you very much for your visit! İngiltere'ye seyahat etmek için en uygun dönem hangisidir? \\

\textbf{Hamza-large}: If I could travel anywhere in the world, I would go to Paris" diyor. Yani her yere gidebilirim, dünyanın her köşesinde yaşayabilirmişim gibi hissediyorum kendimi. Çünkü benim için önemli olan yer değil, yolculuk yapmak. Yaşadığım şehirlerden bahsetmek gerekirse, ilk olarak İstanbul'u çok sevdiğimi söylemeliyim. Her ne kadar son yıllarda ülkemizdeki terör olaylarından dolayı turist sayısında azalma olsa da, hala birçok yabancı insan Türkiye'yi ziyaret etmeye devam ediyor ve edecek. \\

\textbf{Hamza-xlarge}: If I could travel anywhere in the world, I would go to a place where I can get all of what I was looking for. Dünyanın herhangi bir yerinde istediğim her şeyi bulabileceğimi düşünüyordum, ama aradığımı bulmak için her yere bakmam gerektiğini fark ettim. \\

\textbf{GPT2-xl}: If I could travel anywhere in the world, I would go to this place." \\
"I would like to go to that place," I said. "I think it would be a great place for a movie." I didn't know if that was true or not, but I had a feeling that it was. \\

\textbf{Hamza$_{\scriptsize GPT2-xl}$}: I could travel anywhere in the world, I would go to the same place and would be able to travel with my friends. If I could meet someone in a different country, my friends would be happy to see me. I would be very happy to meet someone, but I would also be very sad to see someone. I think that if I could have a friend, I'd be very glad to see him/her/him/him. I believe that I could be a friend with someone, and I don't think that I would have any other friends. \\

\textbf{Mistral-7b}: If I could travel anywhere in the world, I would go to Italy. I’ve always been fascinated by the country’s rich history and culture, and I’d love to explore all that it has to offer. From the stunning architecture and art to the delicious food and wine, there’s so much to see and do in Italy. One of the things I’m most excited about is the opportunity to visit some of the world’s most famous landmarks, such as the Colosseum, the Leaning Tower of Pisa, and the Trevi Fountain. \\

\textbf{Hamza$_{\scriptsize Mistral}$}: If I could travel anywhere in the world, I would go to… sorusunun cevabını sizler için araştırdık. İşte dünyanın en güzel 10 şehrinin sıralaması. Dünya üzerinde yaşayan insanların büyük çoğunluğunun en çok merak ettiği sorulardan birisi de "dünyadaki en iyi şehirler hangileridir?" sorusudur. Bu soruya cevap vermek oldukça zor bir iştir. Çünkü herkesin zevki farklıdır. Bazıları doğal güzellikleri olan yerleri tercih ederken, bazı kişiler ise tarihi dokusu olan ve kültürel zenginliği yüksek olan bölgeleri seçebilmektedir.\\
\end{myboxnote}

\begin{commentprompt}
\paragraph{Comment:} The comparison of the outputs of Mistral-7b and Hamza$_{\scriptsize Mistral}$ clearly demonstrates the effects of further pretraining, where their outputs starts with the same English phrase, but Hamza$_{\scriptsize Mistral}$'s output suddenly continues with Turkish. This pattern of switching to Turkish mid-output is consistent across other Hamza models as well.
\end{commentprompt}

\clearpage
\begin{myboxnote}
\textbf{Prompt-8}: \textit{Artificial Intelligence will replace many jobs}
\tcblower 
\footnotesize
\textbf{Hamza-small}: Artificial Intelligence will replace many jobs in Turkey. \\

\textbf{Hamza-medium}: Artificial Intelligence will replace many jobs in the field of information technology. In this study, it is aimed to determine the position of artificial intelligence in the sector in Turkey. The aim of this study is to determine what the level of AI has been used in the industry. \\

\textbf{Hamza-large}: Artificial Intelligence will replace many jobs with you to make your life easier and easier. \\

\textbf{Hamza-xlarge}: Artificial Intelligence will replace many jobs and jobs will be more important than ever in the future. Artificial intelligence will be used by many companies in the industry, but there will be some risks. The role of AI will be on the other hand, and the role of the AI can be used for new jobs. \\

\textbf{GPT2-xl}: Artificial Intelligence will replace many jobs in the coming decades, and it will change the way we live and work. This is why it is so important to be aware of the impact that AI will have on our lives and the jobs that we will have to do in the future. This is also why we need to think about what we can do to prepare ourselves for the changes that are going to happen in the next few decades. \\

\textbf{Hamza$_{\scriptsize GPT2-xl}$}: Artificial Intelligence will replace many jobs in the world, but they won't necessarily be jobs that are currently being done. \\

\textbf{Mistral-7b}: Artificial Intelligence will replace many jobs in the future, but it will also create new ones. Artificial Intelligence (AI) has been a hot topic in recent years, and for good reason. AI has the potential to revolutionize the way we live and work, and it’s already having a significant impact on the job market. \\

\textbf{Hamza$_{\scriptsize Mistral}$}: Artificial Intelligence will replace many jobs in the next 10-15 years. This is an inevitable process. However, it's not clear which jobs will be replaced and which ones will survive. It's also not clear how long it will take. AI will be able to do many things that humans can't do. For example, AI can see, hear, smell, taste, and touch. AI can also learn and adapt to new situations. \\

\end{myboxnote}

\begin{commentprompt}
\paragraph{Comment:} In this example, all models successfully maintained the English prompt without switching to Turkish midway. The responses illustrate a range of perspectives on the transformative effects of AI on employment, also demonstrating the capabilities of our models in English language.
\end{commentprompt}

\newpage
\section{Turkish Instruction-Tuning (IT) Dataset Creation Prompts}
\label{sec: it-prompt}

\begin{myboxnote}
\textit{text-davinci3 prompt to create the Turkish Instruction-Tuning (IT) dataset.}
\tcblower 
İyi bir dil modeli olan senden 20 fark task seti üretmen bekleniyor. Bu task instructionları daha sonra bir GPT modeline verilecek ve biz bu GPT modelinin verilen instructionları ne kadar iyi olduğunu  degerlendireceğiz. \\

Kurallar ve koşullar aşağıdaki gibi olmalı: \\
\textbf{1.}\hspace*{0.3cm}Instructionlar Türkçe olmalıdır. \\
\textbf{2.}\hspace*{0.3cm}Instruction çeşitliliğini en üst düzeye çıkarmak için fiil tekrarlarından kaçınmalısın. \\
\textbf{3.}\hspace*{0.3cm}Instruction için kullanılan dil çeşitlilik göstermelidir. Örneğin, sorulaı zorunlu instructionlarla birleştirmelisiniz.\\
\textbf{4.}\hspace*{0.3cm}Instructionların türü farklı olmalıdır. Liste; oluşturma, sınıflandırma, düzenleme vb. gibi çeşitli işlem türlerini içermelidir. \\
\textbf{5.}\hspace*{0.3cm}Bir GPT dil modeli, verilen instructionları tamamlayabilmelidir. Örneğin, asistandan herhangi bir görsel veya işitsel çıktı oluşturmasını istememelisin. Başka bir örnek olarak, herhangi bir action yapamadığı için asistandan seni saat 17:00'de uyandırmasını veya hatırlatıcı ayarlamasını istememelisin. \\
\textbf{6.}\hspace*{0.3cm}Instructionlar 1 ila 2 cümle uzunluğunda olmalı. Bir emir cümlesi veya bir soru cümlesi kullanabilirsin. \\
\textbf{7.}\hspace*{0.3cm}Instruction için uygun bir input oluşturmalısın. Input alanı, instruction için sağlanan belirli bir örneği içermelidir. Gerçekçi veriler içermelidirler. Input, öğretimi zorlaştırmak için kaliteli ve önemli bir içerik sağlamalı, ancak ideal olarak 100 kelimeyi geçmemelidir. \\
\textbf{8.}\hspace*{0.3cm}Tüm instructionlar input gerektirmez. Örneğin, bir instruction "dünyanın en yüksek zirvesi nedir" gibi genel bilgiler sorduğunda, belirli bir bağlam sağlamak gerekli değildir. Bu durumda, input alanına basitçe "$<$noinput$>$" koymalısın. \\
\textbf{9.}\hspace*{0.3cm}Output, instruction ve inputa uygun bir yanıt olmalıdır. Outputun 100 kelimeden az olduğundan emin olmalısın. \\
\textbf{10.}\hspace*{0.3cm}En sonunda, toplamda 20 farklı task setimiz olmalı. \\

20 Task listesi: \\
\underline{\textbf{1. Task}} \\
\textbf{\textit{Instruction:}} Yumurta içermeyen, protein içeren ve yaklaşık 700-1000 kalorilik bir kahvaltıda yiyebileceğim bir şey var mı? \\
\textbf{\textit{Input:}} $<$noinput$>$ \\
\textbf{\textit{Output:}} Evet, 1 yulaf ezmeli muzlu protein shake ve 4 şerit domuz pastırması alabilirsiniz. Yulaf ezmeli muzlu protein karışımı, 1/2 su bardağı yulaf ezmesi, 60 gram peynir altı suyu protein tozu, 1/2 orta boy muz, 1 yemek kaşığı keten tohumu yağı ve 1/2 su bardağı su içerebilir ve toplamda yaklaşık 550 kaloridir. 4 serit domuz pastırması yaklaşık 200 kalori içerir. \\
\underline{\textbf{2. Task}} \\
\textbf{\textit{Instruction:}} Verilen çiftler arasındaki ilişki nedir? \\
\textbf{\textit{Input:}} Gece-Gündüz, Sağ-Sol \\
\textbf{\textit{Output:}} Verilen çiftler arasındaki ilişki, zıt olmalarıdır. \\
\underline{\textbf{3. Task}} \\
\textbf{\textit{Instruction:}} Aşağıdaki kişilerin her biri için tek cümlelik bir açıklama oluşturun.
\textbf{\textit{Input:}} Barack Obama, Elon Musk, Taylor Swift.\\
\textbf{\textit{Output:}} Barack Obama, 2009'dan 2017'ye kadar Amerika Birleşik Devletleri'nin 44. başkanı olarak görev yapan Amerikalı bir politikacıdır. Elon Musk, SpaceX'in kurucusu, CEO'su ve baş mühendisidir; ayrıca Tesla'nın CEO'su; Neuralink ve OpenAI'nin kurucu ortağı; Musk Vakfı başkanı; Twitter'ın sahibi ve CEO'sudur. Taylor Alison Swift, Amerikalı şarkıcı ve söz yazarıdır.
\end{myboxnote}



\end{document}